\pdfoutput=1

\documentclass[11pt]{article}

\usepackage{acl}

\usepackage{bm}
\usepackage{multirow}
\usepackage{xcolor}
\usepackage{colortbl}

\usepackage{makecell}
\usepackage{booktabs}
\usepackage{amssymb}
\usepackage{amsmath}
\usepackage{float}
\usepackage{times}  
\usepackage{helvet}  
\usepackage{courier}  
\usepackage{graphicx} 
\usepackage{natbib}  
\usepackage{caption} 
\usepackage{subfigure}

\usepackage{url}

\usepackage{hyperref}

\usepackage[T1]{fontenc}

\usepackage[utf8]{inputenc}

\usepackage{microtype}

\usepackage{inconsolata}

\title{BELLE: A Bi-Level Multi-Agent Reasoning Framework for Multi-Hop Question Answering}

\author{Taolin Zhang$^{1}$, Dongyang Li$^{2}$, Qizhou Chen$^{3,4}$, Chengyu Wang$^{3}$\thanks{\ \ Corresponding author.}, \textbf{Xiaofeng He}$^{4}$\\
$^1$ School of Computer Science and Information Engineering, Hefei University of Technology \\
$^2$ Shanghai University of Electric Power
$^3$ Alibaba Group
$^4$ East China Normal University\\
 {\tt {tlzhang}@hfut.edu.cn, chengyu.wcy@alibaba-inc.com} \\
 }
 
\begin{document}
\maketitle

\begin{abstract}
Multi-hop question answering (QA) involves finding multiple relevant passages and performing step-by-step reasoning to answer complex questions. Previous works on multi-hop QA employ specific methods from different modeling perspectives based on large language models (LLMs), regardless of question types.
In this paper, we first conduct an in-depth analysis of public multi-hop QA benchmarks, categorizing questions into four types and evaluating five types of cutting-edge methods: Chain-of-Thought (CoT), Single-step, Iterative-step, Sub-step, and Adaptive-step. We find that different types of multi-hop questions exhibit varying degrees of sensitivity to different types of methods. 
Thus, we propose a Bi-levEL muLti-agEnt reasoning (BELLE) framework to address multi-hop QA by specifically focusing on the correspondence between question types and methods, with each type of method regarded as an ``operator'' by prompting LLMs differently. The first level of BELLE includes multiple agents that debate to formulate an executable plan of combined ``operators'' to address the multi-hop QA task comprehensively. During the debate, in addition to the basic roles of affirmative debater, negative debater, and judge, at the second level, we further leverage fast and slow debaters to monitor whether changes in viewpoints are reasonable.
Extensive experiments demonstrate that BELLE significantly outperforms strong baselines in various datasets. Additionally, the model consumption of BELLE is higher cost-effectiveness than that of single models in more complex multi-hop QA scenarios.
\end{abstract}

\begin{figure*}[!t]
  \centering
  \includegraphics[height=7cm,width=15cm]{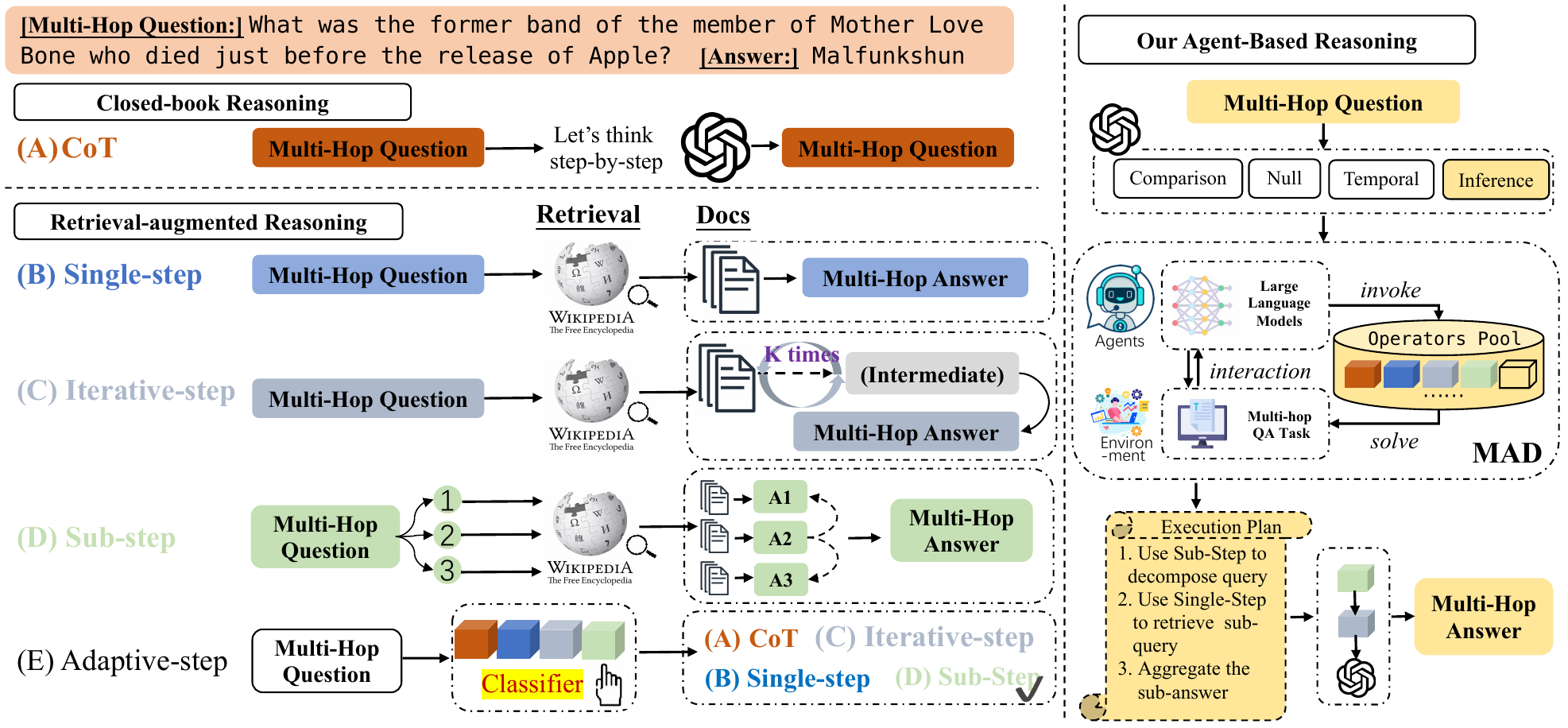}
  \caption{Comparison between our approach and existing methods for multi-hop QA. (1) \texttt{Closed-book reasoning} does not consider the requirement for external knowledge. (2) \texttt{Retrieval-augmented reasoning} leverages an end-to-end fixed solution to solve all multi-hop questions. (3) \texttt{Our agent-based reasoning} framework provides an execution plan to dynamically combine appropriate multi-hop operators with respect to multi-hop question types.}
  \label{motivation_res}
\end{figure*}

\section{Introduction}
\label{introduction}
Recently, large language models (LLMs) have become the fundamental infrastructure of modern NLP~\cite{DBLP:conf/acl/BlevinsGZ23,DBLP:conf/ecai/ZhangLC0H0H024,DBLP:conf/acl/ZhangCL0HHXH24,DBLP:conf/acl/ChuCCWZDYLQ24}. Furthermore, chain-of-thought (CoT) prompting enhances the reasoning capabilities of LLMs~\cite{DBLP:conf/nips/Wei0SBIXCLZ22,DBLP:conf/acl/Shaikh0HBY23,DBLP:conf/acl/ChuCCYH0P00L24}. Yet, the complexity of multi-hop question answering (QA) often surpasses the knowledge boundaries of LLMs, which can lead to factual errors in generated responses, also known as hallucinations~\cite{DBLP:conf/acl/KhalifaLLL023,Huang_2024,DBLP:conf/acl/ChuCCWZDYLQ24,DBLP:conf/acl/Shi00GRCR24}.

In the literature, multi-hop QA approaches with LLMs can be divided into two categories:
\texttt{(1) Closed-book Reasoning:} This approach utilizes the understanding ability of LLMs for multi-hop questions, obtaining refined answers through probabilistic sampling in LLMs' response generation. CoT~\cite{DBLP:conf/nips/Wei0SBIXCLZ22} prompts LLMs step by step for multi-hop questions to generate the reasoning process. Considering complex multi-hop reasoning paths, several works \cite{DBLP:conf/emnlp/DuaG0G22,DBLP:conf/iclr/ZhouSHWS0SCBLC23} decompose them into sub-step questions and solve them progressively, while others \cite{DBLP:conf/nips/YaoYZS00N23,DBLP:conf/acl/ChuCCWZDYLQ24,DBLP:conf/emnlp/MenonZV24} model reasoning procedures as BFS or DFS search on probabilistic reasoning trees. As reported in~\cite{DBLP:conf/icml/BorgeaudMHCRM0L22}, the knowledge learned by LLMs is often insufficient to answer complex questions, which require external data support.
\texttt{(2) Retrieval-augmented Reasoning:} Early work utilizes single-step retrieval, but often struggles to gather all necessary knowledge to answer multi-hop questions, resulting in knowledge omissions~\cite{DBLP:journals/corr/abs-2203-05115,DBLP:conf/icml/BorgeaudMHCRM0L22,DBLP:journals/jmlr/IzacardLLHPSDJRG23}. Several approaches leverage iterative-step retrievals by concatenating output from previous rounds with sub-step questions~\cite{DBLP:conf/emnlp/PressZMSSL23,DBLP:conf/emnlp/ShaoGSHDC23,DBLP:journals/corr/abs-2407-13101}. As shown in Fig.~\ref{motivation_res}, no matter what multi-hop question is given, retrieval methods directly recall external knowledge and answer the question with integrated inputs. Although the adaptive-step method leverages classifiers for different questions~\cite{DBLP:conf/naacl/JeongBCHP24}, they still use a fixed approach, regardless of question types. This also incurs an additional computational burden for relatively simple questions, which limits their usage in applications that require high inference speed~\cite{DBLP:journals/ftir/MaviJJ24,DBLP:conf/emnlp/ZhuangZCYLHLR0Z24}.

\begin{figure*}[!t]
  \centering
  \includegraphics[width=16cm,height=5.5cm]{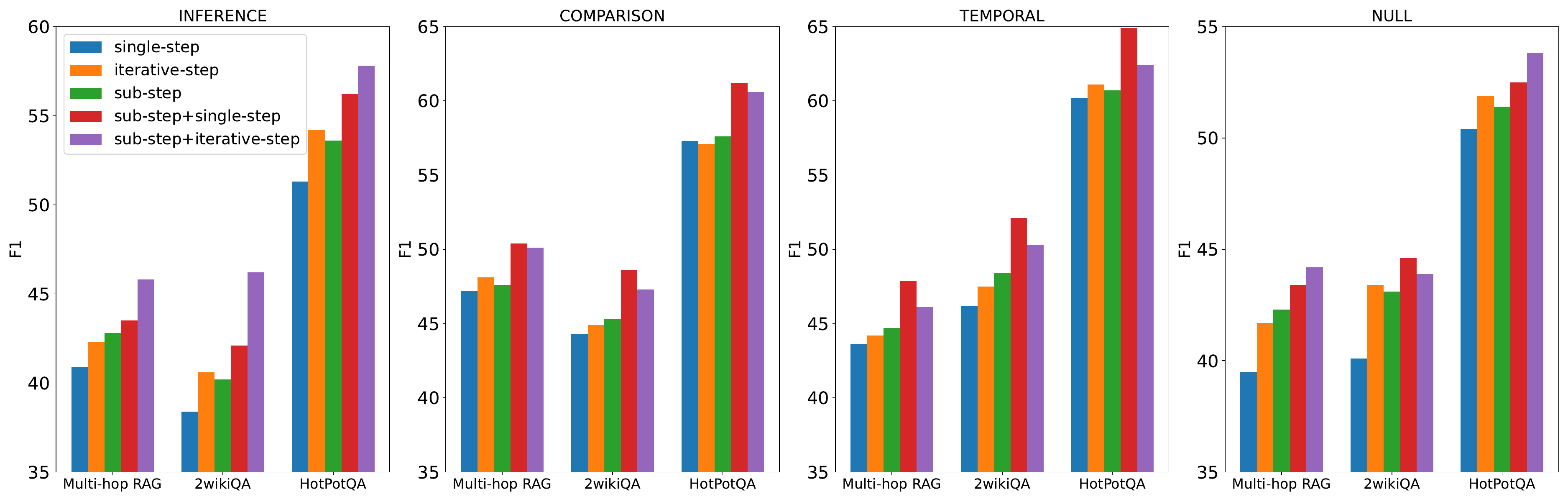}
  \caption{Comparison of single and combined operators in different multi-hop question types. The red and purple bars represent the combined operators of sub-step + single-step and sub-step + iterative-step, respectively.}
  \label{data_analysis}
\end{figure*}

To overcome the above problems, our research focuses on the following question: \textit{How can we dynamically combine various operators based on question types to improve the performance of multi-hop QA, while reducing the computational overhead?} Building on this motivation, we present a novel bi-level multi-agent system named BELLE, which creates and executes a plan of operators\footnote{We view specific solutions (e.g., CoT~\cite{DBLP:conf/nips/Wei0SBIXCLZ22}) as ``operators'' from the perspective of prompting LLMs.} for answering multi-hop questions where the plan is represented by the output summary of our multi-agent debate (MAD) system.

Specifically, we first conduct an analysis on whether different types of multi-hop questions are better answered by different operators. Following \cite{DBLP:journals/corr/abs-2401-15391}, the four question types are \texttt{Inference}, \texttt{Comparison}, \texttt{Temporal}, and \texttt{Null}. From Fig.~\ref{data_analysis}, the \texttt{Temporal} and \texttt{Comparison} types are relatively simple, requiring only breaking down the question into sub-questions and using a single-step retrieval method to recall the fact. However, for the \texttt{Inference} type, due to their complexity, it is necessary to break down the question and use iterative-step retrieval to obtain more external knowledge. For other questions, we can directly use the LLM's internal knowledge to answer them.

Based on the analysis, the multi-agent pipeline consists of three modules.
(i) \texttt{Question Type Classification:} We provide in-context examples formatted as new QA pairs, and inputs to LLMs are classified into the four question types.
(ii) \texttt{Bi-Level Multi-agent Debate:} In addition to the basic roles in multi-agent systems~\cite{DBLP:conf/emnlp/LiTW024,DBLP:conf/emnlp/Liang0JW00Y0T24}, we propose a bi-level architecture including a slow-debater and a fast-debater to fully utilize both the historical discussion and the current state of opposing sides to determine which multi-hop QA operators to invoke~\cite{DBLP:journals/corr/abs-2410-08328}. Our objective is to maximize the use of information already discussed for planning operators while also preventing bias in the agent's viewpoint~\cite{DBLP:conf/emnlp/TaubenfeldDRG24,DBLP:conf/emnlp/BorahM24}.
(iii) \texttt{Multi-hop QA Executor:} When the system provides a plan to invoke specific operators, we use LLMs again to generate responses according to the plan. Finally, we concatenate the results of each step to obtain sub-answers and trace back to the root node to achieve the final answer for the multi-hop question.

We evaluate BELLE on four multi-hop QA datasets, including MultiHop-RAG~\cite{DBLP:journals/corr/abs-2401-15391}, 2WikiMultiHopQA~\cite{DBLP:conf/coling/HoNSA20}, HotPotQA~\cite{DBLP:conf/emnlp/Yang0ZBCSM18}, and MuSiQue~\cite{DBLP:journals/tacl/TrivediBKS22}.
The experiments are conducted using GPT-3.5-turbo~\cite{DBLP:conf/nips/BrownMRSKDNSSAA20} and Qwen2.5-7B~\cite{qwen25}.
The results show that our method significantly outperforms baselines. An analysis on more difficult multi-hop questions reveals the computational cost superiority of our dynamic operators combination.


\section{Related Works}

\noindent \textbf{Multi-Hop Question Answering.} Multi-hop QA is more complex than simple QA because it involves not just retrieving information, but also effectively combining related facts. Facts can be sourced from a knowledge graph~\cite{DBLP:conf/emnlp/LinSX18,DBLP:conf/icpads/ChengNMC23,DBLP:conf/emnlp/ZhongWMPC23}, tables~\cite{DBLP:journals/debu/LuLK16}, free-form text~\cite{DBLP:conf/emnlp/Yang0ZBCSM18,DBLP:journals/tacl/WelblSR18}, or a heterogeneous combination of these sources~\cite{DBLP:conf/emnlp/ChenZCXWW20,DBLP:journals/corr/abs-2204-09140,DBLP:conf/acl/LeiLWHHZL23}. With the development of LLMs, prompt-based methods combined with an optional retrieval module have become a popular approach for handling multi-hop QA~\cite{DBLP:conf/emnlp/PressZMSSL23,DBLP:conf/emnlp/ZhongWMPC23,DBLP:conf/emnlp/ZhuangZCYLHLR0Z24,DBLP:conf/acl/ChuCCWZDYLQ24}.
Recently, the agent-based methods for multi-hop QA are also proposed \cite{DBLP:journals/corr/abs-2412-18431,wu2025talkrightspecialistsrouting}.
While all previous works focus on a specific multi-hop QA method, our approach targets a dynamic, flexible pipeline from a more fine-grained question type perspective.

\noindent \textbf{Multi-Agent Debate of LLMs.} Current approaches to multi-agent debate (MAD) can generally be divided into two main categories: (1) Those that adjust the model prompts and responses during the debate~\cite{DBLP:conf/emnlp/Liang0JW00Y0T24,DBLP:conf/icml/KhanHVRSRGBRP24,DBLP:journals/corr/abs-2401-01312,DBLP:conf/acl/FengS00BT24,DBLP:journals/corr/abs-2404-09127}. These MAD methods generate specific opinions in response to particular situations while solving a task. (2) Those that alter the structure of the debate process~\cite{DBLP:journals/corr/abs-2303-17760,DBLP:journals/corr/abs-2310-02170,DBLP:journals/corr/abs-2402-06634,DBLP:conf/iclr/HongZCZCWZWYLZR24}. Importantly, both categories use off-the-shelf LLMs (e.g., API) and work by modifying either the inputs or outputs of these models. However, previous work did not take into account the comprehensive utilization of historical and current information in multi-agent collaboration, resulting in a waste of information.

\section{Analysis of Multi-Hop Question Types}
\label{analysis_mhq}

In this section, we analyze the sensitivity of different types of multi-hop questions involving single and combined operators as described previously.

We leverage four multi-hop QA datasets, namely MultiHop-RAG~\cite{DBLP:journals/corr/abs-2401-15391}, 2WikiMultiHopQA~\cite{DBLP:conf/coling/HoNSA20}, HotPotQA~\cite{DBLP:conf/emnlp/Yang0ZBCSM18}, and MuSiQue~\cite{DBLP:journals/tacl/TrivediBKS22} as the data sources.\footnote{The complete results and the analysis of the question type annotation process are shown in Appendix \ref{overall_results_da}.} The other three datasets, except for MultiHop-RAG, do not include question type labels. Hence, we use GPT-4~\cite{gpt4_} to annotate half of the datasets and perform cross-validation. The prompt for label annotation is shown in Appendix~\ref{question_type_annotate}. Considering potential annotation errors by LLMs, we refine the prompts and manually check the responses to select suitable prompts. During the manual verification of data labeling, two individuals independently test 100 samples of each type. A prompt is adopted only if both individuals agree that the labeling is consistent with the actual question type, achieving an accuracy of 95\%. To maintain consistency in the label space,\footnote{Due to the extensibility of our BELLE, there will be more fine-grained question type classification rules that can be directly used by modifying the \texttt{Meta Prompt} in the future.} we set it to be the same as that of MultiHop-RAG, which includes four types: \texttt{Inference}, \texttt{Comparison}, \texttt{Temporal}, and \texttt{Null}.

As for the combined operators, we have selected two representative methods: sub-step+single-step and sub-step+iterative-step. From Fig.~\ref{data_analysis}, we can draw two conclusions:

\noindent\textbf{1.~Combined operators are superior to single operators in multi-hop QA tasks.} Across the four question types, the method of combined operators consistently outperforms single operators. On average, the performance of combined operators is 3\% higher than that of single operators across different question types and datasets.

\noindent\textbf{2.~Different combinations of operators have varying degrees of sensitivity to question types.} For the \texttt{Inference} type, due to the increase in logical reasoning steps, it is necessary to recall more external knowledge~\cite{DBLP:journals/corr/abs-2204-09140,DBLP:journals/ftir/MaviJJ24}. In this case, decomposing the complex question and combining it with a multi-round retrieval scheme is more suitable for this multi-hop question type. For \texttt{Comparison} and \texttt{Temporal} types, we typically only need to identify the important subjects (e.g., entity or timestamp) for these question types and retrieve relevant content. Hence, the method based on sub-questions combined with single-step retrieval can address them effectively.

Therefore, using different combinations of operators is better for solving different types of multi-hop questions than using a specific operator alone.



\begin{figure*}[!t]
  \centering
  \includegraphics[height=8cm,width=15cm]{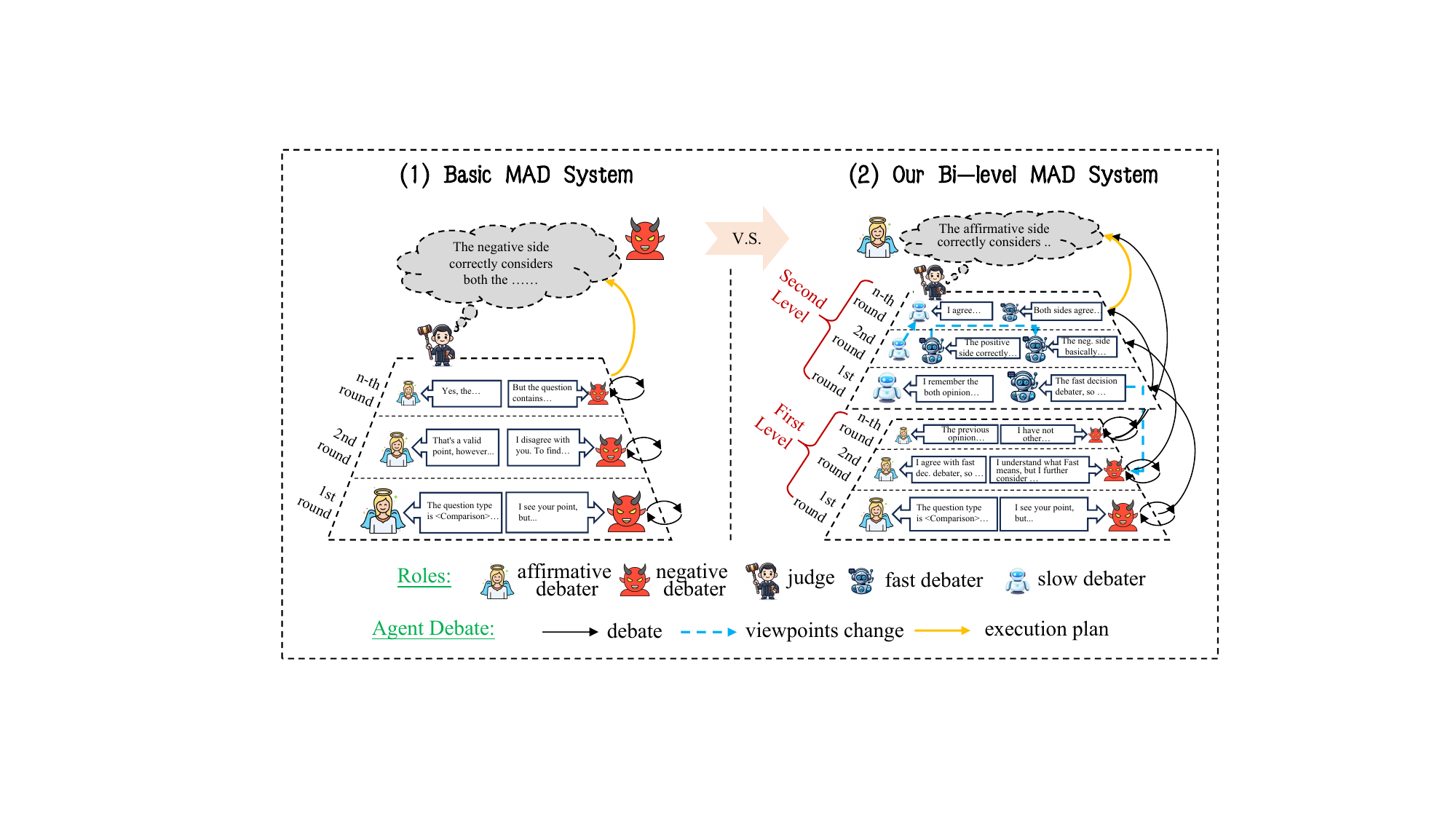}
  \caption{Model overview of BELLE. The left part is the existing MAD system containing three basic roles (i.e., an affirmative side, a negative side and a judge). The right part is the details of our bi-level MAD system including first-level and second-level debaters.}
  \label{model_arch}
\end{figure*}


\section{Methodology}
\label{methodology}
In this section, we provide a detailed description of BELLE, with the bi-level MAD system shown in Fig.~\ref{model_arch}. Our framework includes the following three modules:
(i) \texttt{Question Type Classifier:} Multi-hop questions are classified into the corresponding question types as discussed in Sect.~\ref{analysis_mhq}.
(ii) \texttt{Bi-Level Multi-agent Debater:} In addition to conventional MAD systems, slow and fast debaters are proposed to aid opposing sides in invoking the operators with historical discussion.
(iii) \texttt{Multi-hop QA Executor:} It executes the planning of operators to answer multi-hop questions.

\subsection{Question Type Classifier}
\label{question_type_classifier}
Compared to previous works~\cite{DBLP:conf/icpads/ChengNMC23,DBLP:conf/acl/ChuCCWZDYLQ24,DBLP:conf/emnlp/ZhuangZCYLHLR0Z24} that use a specific method to coarsely solve multi-hop QA tasks, we find that the complex multi-hop reasoning task requires dynamic combinations of operators based on question types. Hence, BELLE first considers fine-grained classification of multi-hop questions as input for subsequent modules.

Specifically, this module can be directly formalized as a text classification task, denoted as $\mathcal{A}_{t} = \mathcal{M}_{t}(q)$. Here, $q$ denotes a multi-hop question, and $\mathcal{M}_t$ is the LLM for question type classification. As for $\mathcal{A}_t$, we use the four question types analyzed in Sect.~\ref{analysis_mhq} as the output label space. We concatenate several QA examples as demonstrations to perform the ICL mechanism,\footnote{Other mechanisms are also analyzed in Appendix \ref{impact_type_classifier}.} ensuring output of the correct question type labels and preventing the instruction degradation phenomenon~\cite{DBLP:conf/nips/BrownMRSKDNSSAA20,DBLP:journals/corr/abs-2410-15553}. The detailed format of templates is described in Appendix~\ref{question_type_annotate}.

\subsection{Bi-Level Multi-agent Debate}
\label{bilevel_mad}
Recently, many MAD systems have addressed specific scenarios with a setting consisting of an affirmative debater, a negative debater, and a judge~\cite{DBLP:conf/emnlp/HeC0LLSZ23,DBLP:conf/emnlp/LiTW024}. These agents can only make a decision for task solutions based on the current state, while the historical discussion contents are not fully utilized. Consequently, the task viewpoints of both debaters may be uncontrollably altered due to the influence of one another~\cite{DBLP:conf/emnlp/TaubenfeldDRG24,DBLP:conf/emnlp/BorahM24}.

Inspired by \citet{DBLP:journals/corr/abs-2410-08328}, we introduce a bi-level MAD system, which employs two additional memory agents named slow-debater and fast-debater to integrate the relationship between historical discussions and current viewpoints. Next, we provide a detailed description of our system, where two representative opposing debaters, two memory debaters, and a judge are involved in a debate to resolve a multi-hop question. Our framework is composed of four components divided into two levels, elaborated as follows.

\subsubsection{The First Level of Debate}
\label{first_level_debate}

\noindent\textbf{Meta Prompts and Operators.} Considering that agents initially might not understand the task, we leverage meta prompts to introduce the question type $\mathcal{A}_{t}$, the number of debaters, the round limit, and other requirements, as shown in Appendix~\ref{meta_prompts}. We create an atmosphere for debaters to engage in a "tit for tat" debate (see indicated contents).

For the operators pool, each element will be invoked by the following bi-level MAD system, selecting from two paradigms described in Fig.~\ref{motivation_res}. We choose CoT~\cite{DBLP:conf/nips/Wei0SBIXCLZ22}, single-step~\cite{DBLP:journals/jmlr/IzacardLLHPSDJRG23}, iterative-step~\cite{DBLP:conf/acl/TrivediBKS23}, sub-step question~\cite{DBLP:conf/emnlp/PressZMSSL23}, and adaptive-step~\cite{DBLP:conf/naacl/JeongBCHP24} as representative operators.

\noindent\textbf{Opposing Debaters.} There are two debaters that play the roles of the affirmative and the negative, respectively. In each debate round, the debaters take turns presenting arguments based on their own previous debate history. For the affirmative debater, denote the debate history from all \texttt{t-1} rounds as $H_{ad}^{t-1}$.
The result of the $t$-th round discussion for the affirmative debater is defined as follows:
\begin{equation}
f_{ad}^t = \mathcal{M}(H_{ad}^{t-1}, f_{fast}^{t-1}, f_{slow}^{t-1})
\end{equation}
where $\mathcal{M}$ is the same LLM as $\mathcal{M}_{t}$.
$f_{fast}^{t-1}$ and $f_{slow}^{t-1}$ represent the discussion results of the fast and slow debaters in the \texttt{(t-1)}-th round, respectively.
The definitions for the debate history and discussion results of the negative debater, denoted as $f_{nd}^t$, are similarly defined.

\subsubsection{The Second Level of Debate}

The first level of discussion focuses on each side's positions without evaluating the rationality of operator selection. Therefore, in our proposed bi-level debate mechanism, the second level comprehensively evaluates the operator selection in the current $t$-th round (fast debater) and summarizes historical debates (slow debater).

\noindent\textbf{Fast Debater.} In the discussion process of the fast debater, the main goal is to assess whether the operators selected in the current discussion between both sides are reasonable. This involves the participation of three roles: the affirmative and negative sides in the $t$-th round, as well as the previous discussion results of the fast debater. We denote the debate history of the fast debater from all previous \texttt{t-1} rounds as $H_{fast}^{t-1}$.
Hence, the current $t$-th debate result of the fast debater is as follows:
\begin{equation}
    f_{fast}^t = \mathcal{M}(f_{ad}^t, f_{nd}^t, H_{fast}^{t-1})
\end{equation}
Note that the fast debater only considers the situation in the current $t$-th debate, making it susceptible to the viewpoints of both sides, as illustrated by the blue dashed line in Fig.~\ref{model_arch}.

\noindent\textbf{Slow Debater.} Compared to the fast debater, the slow debater integrates all historical information to judge the rationality of operator selection. The more important goal is to prevent debaters from losing confidence in correct viewpoints, which may lead to oscillation~\cite{DBLP:journals/corr/abs-2311-13884}. The slow debater process involves the affirmative, negative, fast, and historical roles of the slow debater. Similar to the fast debater, the debate history from all previous \texttt{t-1} rounds is $H_{slow}^{t-1}$. The current viewpoint of the slow debater is as follows:
\begin{equation}
f_{slow}^t = \mathcal{M}(f_{ad}^t, f_{nd}^t, f_{fast}^{t}, H_{slow}^{t-1})
\end{equation}

\noindent\textbf{Judge.} Finally, we design a judge $J$ to oversee the debate process, providing an execution plan of combined operators. The judge operates in two modes: (a) Hard Mode, where judge $J$ decides if a correct combination of operators can be determined after all debaters present their viewpoints. If possible, the debate concludes; otherwise, it continues. (b) Soft Mode, where judge $J$ extracts useful operator suggestions based on the slow debater's history, $H_{slow}^{t}$, since no correct solution is found within the debate's round limit. The judge's template is in Appendix~\ref{meta_prompts}, which produces a summarized plan for invoking operators step by step.

\subsection{Multi-hop QA Executor}
Through the discussion of our bi-level MAD system, we have obtained the specific plan for solving the multi-hop question. Then, we progressively invoke the corresponding multi-hop operators to obtain the final answer. To ensure consistency in the LLM's understanding, we use the same LLM $\mathcal{M}$ to execute the sub-steps of the operator planning process. An example is shown in Appendix~\ref{examples_operators}.

\begin{table*}[t]
\centering
\begin{small}
\begin{tabular}{ccccccccccccc}
\toprule
 \multicolumn{1}{c|}{\multirow{1}{*}{Dataset$ \rightarrow$}} & \multicolumn{3}{c|}{Multi-hop RAG}                                                & \multicolumn{3}{c|}{HotpotQA}                                                     & \multicolumn{3}{c|}{2WikiQA}  & \multicolumn{3}{c}{MuSiQue}                                                     \\ \cmidrule{2-13} 
 \multicolumn{1}{c|}{\multirow{1}{*}{Models$ \downarrow$}}                       & \multicolumn{1}{c|}{EM}   & \multicolumn{1}{c|}{F1}   & \multicolumn{1}{c|}{Acc}  & \multicolumn{1}{c|}{EM}   & \multicolumn{1}{c|}{F1}   & \multicolumn{1}{c|}{Acc}  & \multicolumn{1}{c|}{EM}   & \multicolumn{1}{c|}{F1}    & \multicolumn{1}{c|}{Acc} & \multicolumn{1}{c|}{2hop}   & \multicolumn{1}{c|}{3hop}    & \multicolumn{1}{c}{4hop} \\ \midrule
\rowcolor[HTML]{C0C0C0}\multicolumn{13}{c}{Closed-book Reasoning}    \\ \midrule
\multicolumn{1}{c|}{SP}                       & \multicolumn{1}{c|}{39.4} & \multicolumn{1}{c|}{47.5} & \multicolumn{1}{c|}{44.3} & \multicolumn{1}{c|}{32.1} & \multicolumn{1}{c|}{38.9} & \multicolumn{1}{c|}{37.4} & \multicolumn{1}{c|}{27.8} & \multicolumn{1}{c|}{33.9}  & \multicolumn{1}{c|}{31.6}  & 
\multicolumn{1}{c|}{16.4} & \multicolumn{1}{c|}{16.2} & \multicolumn{1}{c}{12.6}  \\ 

\multicolumn{1}{c|}{CoT}                      & \multicolumn{1}{c|}{43.6} & \multicolumn{1}{c|}{50.5} & \multicolumn{1}{c|}{49.7} & \multicolumn{1}{c|}{40.5} & \multicolumn{1}{c|}{46.5} & \multicolumn{1}{c|}{47.3} & \multicolumn{1}{c|}{36.2} & \multicolumn{1}{c|}{42.3}  & \multicolumn{1}{c|}{43.7}    & \multicolumn{1}{c|}{30.2} & \multicolumn{1}{c|}{22.5}  & \multicolumn{1}{c}{13.2}  \\ \midrule

\rowcolor[HTML]{C0C0C0}\multicolumn{13}{c}{Retrieval-augmented Reasoning}      \\ \midrule
\multicolumn{1}{c|}{Single-step}             & \multicolumn{1}{c|}{47.2} & \multicolumn{1}{c|}{52.3} & \multicolumn{1}{c|}{51.3} & \multicolumn{1}{c|}{48.7} & \multicolumn{1}{c|}{55.3} & \multicolumn{1}{c|}{54.6} & \multicolumn{1}{c|}{38.1} & \multicolumn{1}{c|}{42.9}  & \multicolumn{1}{c|}{41.3}    & \multicolumn{1}{c|}{22.1} & \multicolumn{1}{c|}{10.6}  & \multicolumn{1}{c}{10.4}     \\ 

\multicolumn{1}{c|}{Self-Ask}                 & \multicolumn{1}{c|}{49.8} & \multicolumn{1}{c|}{54.6} & \multicolumn{1}{c|}{52.6} & \multicolumn{1}{c|}{44.5} & \multicolumn{1}{c|}{49.4} & \multicolumn{1}{c|}{50.2} & \multicolumn{1}{c|}{40.5} & \multicolumn{1}{c|}{46.9}  & \multicolumn{1}{c|}{48.5}      & \multicolumn{1}{c|}{24.4} & \multicolumn{1}{c|}{8.8}  & \multicolumn{1}{c}{7.5 }   \\

\multicolumn{1}{c|}{IRCoT}                    & \multicolumn{1}{c|}{55.1} & \multicolumn{1}{c|}{59.2} & \multicolumn{1}{c|}{58.4} & \multicolumn{1}{c|}{51.2} & \multicolumn{1}{c|}{56.2} & \multicolumn{1}{c|}{55.4} & \multicolumn{1}{c|}{50.7} & \multicolumn{1}{c|}{56.8}  & \multicolumn{1}{c|}{52.3}  & \multicolumn{1}{c|}{31.4} & \multicolumn{1}{c|}{19.2}  & \multicolumn{1}{c}{16.4}  \\ 

\multicolumn{1}{c|}{FLARE}                    & \multicolumn{1}{c|}{54.9} & \multicolumn{1}{c|}{58.7} & \multicolumn{1}{c|}{59.2} & \multicolumn{1}{c|}{50.8} & \multicolumn{1}{c|}{56.1} & \multicolumn{1}{c|}{58.3} & \multicolumn{1}{c|}{58.2} & \multicolumn{1}{c|}{60.1}  & \multicolumn{1}{c|}{63.7}  & \multicolumn{1}{c|}{40.9} & \multicolumn{1}{c|}{27.1}  & \multicolumn{1}{c}{15.0 }     \\ 

\multicolumn{1}{c|}{ProbTree}                 & \multicolumn{1}{c|}{56.5} & \multicolumn{1}{c|}{62.5} & \multicolumn{1}{c|}{60.1} & \multicolumn{1}{c|}{56.3} & \multicolumn{1}{c|}{60.4} & \multicolumn{1}{c|}{60.6} & \multicolumn{1}{c|}{64.3} & \multicolumn{1}{c|}{67.9}  & \multicolumn{1}{c|}{65.4}      & \multicolumn{1}{c|}{41.2} & \multicolumn{1}{c|}{30.9}  & \multicolumn{1}{c}{14.4}    \\ 

\multicolumn{1}{c|}{EfficientRAG}             & \multicolumn{1}{c|}{49.2} & \multicolumn{1}{c|}{55.3} & \multicolumn{1}{c|}{54.7} & \multicolumn{1}{c|}{52.9} & \multicolumn{1}{c|}{57.9} & \multicolumn{1}{c|}{55.4} & \multicolumn{1}{c|}{47.7} & \multicolumn{1}{c|}{51.6} & \multicolumn{1}{c|}{53.8}   & \multicolumn{1}{c|}{32.7} & \multicolumn{1}{c|}{23.6} & \multicolumn{1}{c}{12.5}     \\

\multicolumn{1}{c|}{BeamAggR}                 & \multicolumn{1}{c|}{61.9} & \multicolumn{1}{c|}{\underline{67.2}} & \multicolumn{1}{c|}{66.8} & \multicolumn{1}{c|}{55.6} & \multicolumn{1}{c|}{\underline{62.9}} & \multicolumn{1}{c|}{59.2} & \multicolumn{1}{c|}{66.1} & \multicolumn{1}{c|}{\underline{71.6}}  & \multicolumn{1}{c|}{69.2}      & \multicolumn{1}{c|}{\underline{45.9}} & \multicolumn{1}{c|}{\underline{36.8}}  & \multicolumn{1}{c}{\underline{21.6}}   \\ \midrule

\rowcolor[HTML]{C0C0C0}\multicolumn{13}{c}{Agent-based Reasoning}      \\ \midrule

\multicolumn{1}{c|}{LONGAGENT}             & \multicolumn{1}{c|}{53.6} & \multicolumn{1}{c|}{56.8} & \multicolumn{1}{c|}{57.4} & \multicolumn{1}{c|}{52.4} & \multicolumn{1}{c|}{59.3} & \multicolumn{1}{c|}{58.1} & \multicolumn{1}{c|}{60.1} & \multicolumn{1}{c|}{65.6} & \multicolumn{1}{c|}{62.8}   & \multicolumn{1}{c|}{40.5} & \multicolumn{1}{c|}{25.8} & \multicolumn{1}{c}{16.4}     \\ 

\multicolumn{1}{c|}{GEAR}             & \multicolumn{1}{c|}{50.7} & \multicolumn{1}{c|}{52.5} & \multicolumn{1}{c|}{51.9} & \multicolumn{1}{c|}{50.4} & \multicolumn{1}{c|}{54.6} & \multicolumn{1}{c|}{54.8} & \multicolumn{1}{c|}{47.4} & \multicolumn{1}{c|}{52.3} & \multicolumn{1}{c|}{51.6}   & \multicolumn{1}{c|}{35.1} & \multicolumn{1}{c|}{20.9} & \multicolumn{1}{c}{15.3}     \\
\multicolumn{1}{c|}{RopMura}             & \multicolumn{1}{c|}{52.6} & \multicolumn{1}{c|}{53.7} & \multicolumn{1}{c|}{58.2} & \multicolumn{1}{c|}{49.2} & \multicolumn{1}{c|}{53.1} & \multicolumn{1}{c|}{55.7} & \multicolumn{1}{c|}{58.8} & \multicolumn{1}{c|}{63.2} & \multicolumn{1}{c|}{64.0}   & \multicolumn{1}{c|}{41.1} & \multicolumn{1}{c|}{24.6} & \multicolumn{1}{c}{16.2}     \\ \midrule
\multicolumn{1}{c|}{BELLE}                    & \multicolumn{1}{c|}{\textbf{64.7}} & \multicolumn{1}{c|}{ \begin{tabular}[c]{@{}c@{}}\textbf{70.4}\\ $(\uparrow3.2)$ \end{tabular} }  & \multicolumn{1}{c|}{\textbf{68.5}} & \multicolumn{1}{c|}{\textbf{59.2}} & \multicolumn{1}{c|}{\begin{tabular}[c]{@{}c@{}}\textbf{66.5}\\ $(\uparrow3.6)$ \end{tabular} } & \multicolumn{1}{c|}{\textbf{63.7}} & \multicolumn{1}{c|}{\textbf{69.7}} & \multicolumn{1}{c|}{\begin{tabular}[c]{@{}c@{}}\textbf{75.7}\\ $(\uparrow4.1)$ \end{tabular} }  & \multicolumn{1}{c|}{\textbf{72.8}}   & \multicolumn{1}{c|}{\begin{tabular}[c]{@{}c@{}}\textbf{50.5}\\ $(\uparrow4.6)$ \end{tabular}} & \multicolumn{1}{c|}{\begin{tabular}[c]{@{}c@{}}\textbf{42.1}\\ $(\uparrow5.3)$ \end{tabular}}  & \multicolumn{1}{c}{\begin{tabular}[c]{@{}c@{}}\textbf{29.2}\\ $(\uparrow7.6)$ \end{tabular}}    \\ \bottomrule
\end{tabular}
\end{small}
\caption{The general results of BELLE. The best and second results are highlighted by \textbf{bold} and \underline{underline}. We show the F1 for 2,3,4-hops of MusiQue. T-tests show the improvements are statistically significant with $p< 0.05$ (\%).}
\label{main_results}
\end{table*}

\section{Experiments}\label{experiments_result}
Due to space limitation, we describe datasets, baselines and implementation details in Appendix \ref{experimental_details}.

\subsection{Experimental Results}

\subsubsection{Results of Multi-hop QA Tasks}

\noindent\textbf{Main Results.} 
Table~\ref{main_results} shows the general performance of BELLE across the four multi-hop QA datasets. We observe that: (1) Generally, due to the requirement for external knowledge in complex multi-hop questions~\cite{DBLP:journals/ftir/MaviJJ24,DBLP:journals/corr/abs-2402-06196}, retrieval-augmented reasoning methods show more significant improvement compared to closed-book methods. However, a comparable improvement can still be achieved by reasoning step by step using CoT~\cite{DBLP:conf/nips/Wei0SBIXCLZ22}. (2) Among retrieval-augmented methods, the simple retrieval method does not significantly improve the effectiveness of multi-hop QA. Other methods with additional enhancement operations, such as ProbTree~\cite{DBLP:conf/emnlp/CaoZSL0THL23} and BeamAggR~\cite{DBLP:conf/acl/ChuCCWZDYLQ24}, achieve significant improvements.
(3) Since the agent-based methods are designed with special modules, the collaborative semantic understanding of multi-hop questions by these methods has not been fully utilized compared to our unified operators' framework.
Therefore, an agent-based approach is still insufficient in solving multi-hop QA tasks.
(4) BELLE consistently achieves the best results. Through careful debate for choosing combined operators, our model achieves the greatest improvement on the extremely difficult MuSiQue dataset under 2, 3, and 4 hops settings.

\begin{figure*}[!t]
  \centering
  \includegraphics[height=5cm, width=16cm]{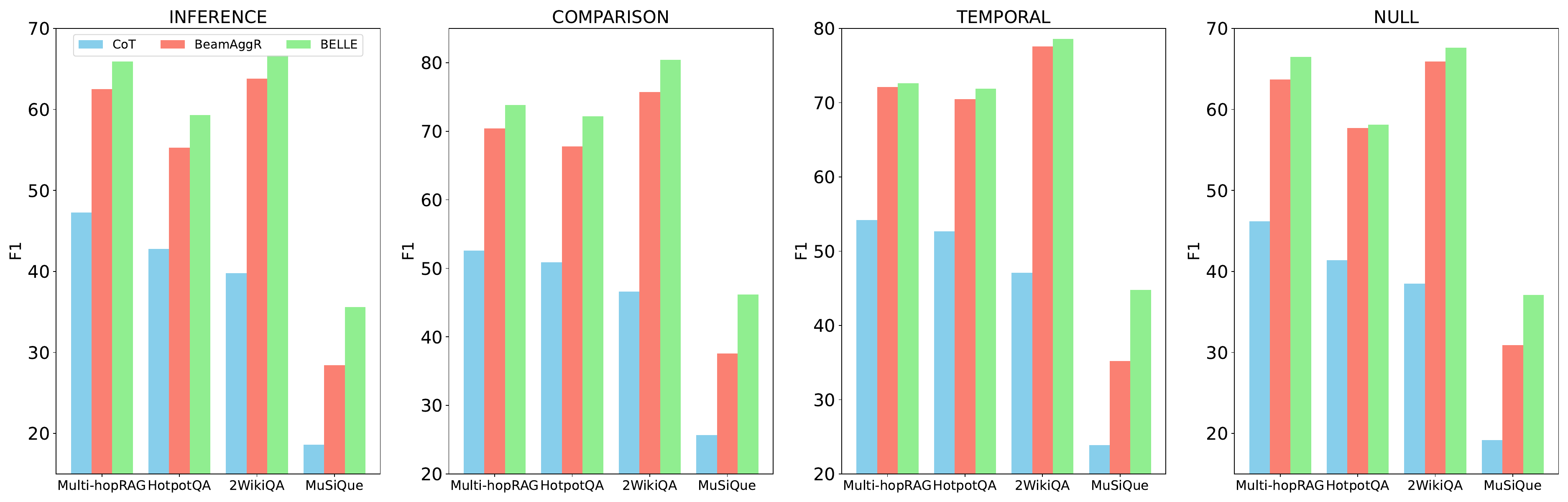}
  \caption{Results of different question types in terms of F1 (\%).}
  \label{diff_type_res}
\end{figure*}

\noindent\textbf{Results of Question Types.}
We present the results for the four types in Fig.~\ref{diff_type_res}, using two strong baselines: CoT~\cite{DBLP:conf/nips/Wei0SBIXCLZ22} and BeamAggR~\cite{DBLP:conf/acl/ChuCCWZDYLQ24}.
Specifically, we observe that (1) The retrieval-based method that introduces external knowledge performs much better on various types of multi-hop questions than simply using an LLM to answer.
Meanwhile, our combined operators method also consistently performs better than the strongest multi-source knowledge-enhanced method.
(2) Our model shows no significant improvement for \texttt{Comparison} and \texttt{Temporal} due to the simple answer patterns.
For \texttt{Comparison} questions, the model only needs to decompose the question into two parts that require comparison, and the answers are concise (e.g., "Yes" and "True").
For \texttt{Temporal} questions, it is usually necessary to find the important timestamp for answering.
However, for the remaining two types, \texttt{Inference} and \texttt{Null}, which are much more difficult, our BELLE model achieves significant improvements.
\texttt{Inference} type questions require reasoning across multiple documents.\footnote{For example, there are two gold paragraphs and eight distractors in HotpotQA~\cite{DBLP:conf/emnlp/Yang0ZBCSM18} for each question.}
Due to the lack of a unified pattern for \texttt{Null} questions, it requires invoking different operators for adaptive combination.

\begin{table}[t]
\centering
\scriptsize
\begin{tabular}{lccccc}
\toprule
Model $\downarrow$ Dataset $\rightarrow$ & \textbf{D1} & \textbf{D2} & \textbf{D3} & \textbf{D4} & \textbf{Avg.} \\
\midrule
\rowcolor[HTML]{C0C0C0}\multicolumn{6}{c}{\texttt{Qwen2.5-7B}}   \\ \midrule
\multicolumn{1}{c|}{BELLE} & 64.1 & 59.4 & 68.5 & 32.8 & 56.2 \\
\multicolumn{1}{c|}{BeamAggR} & 55.8 & 51.8 & 62.4 & 23.2 & 48.3 \\ \midrule
\multicolumn{1}{c|}{w/o Type Classifier} & 59.6 & 54.1 & 63.5 & 25.9 & 50.8 \\
\multicolumn{1}{c|}{w/o First Level Debate} & 61.2 & 55.4 & 64.6 & 28.9 & 52.5 \\
\multicolumn{1}{c|}{w/o Second Level Debate} & 58.8 & 53.5 & 62.1 & 25.4 & 50.0 \\ \midrule
\rowcolor[HTML]{C0C0C0}\multicolumn{6}{c}{\texttt{GPT-3.5-turbo}}   \\ \midrule
\multicolumn{1}{c|}{BELLE} & 70.4 & 66.5 & 75.7 & 40.6 & 63.3 \\ 
\multicolumn{1}{c|}{BeamAggR} & 67.2 & 62.9 & 71.6 & 34.8 & 59.1 \\ \midrule
\multicolumn{1}{c|}{w/o Type Classifier} & 67.9 & 63.4 & 73.2 & 37.6 & 60.5 \\
\multicolumn{1}{c|}{w/o First Level Debate} & 68.2 & 63.7 & 73.5 & 38.1 & 60.9 \\
\multicolumn{1}{c|}{w/o Second Level Debate} & 66.8 & 62.8 & 72.3 & 36.5 & 59.6  \\ \midrule
\multicolumn{1}{c|}{w/o affir.\&neg. Debater} & 68.4 & 64.1 & 73.9 & 38.5 & 61.2 \\
\multicolumn{1}{c|}{w/o Fast Debater} & 67.3 & 63.2 & 72.9 & 37.4 & 60.2 \\
\multicolumn{1}{c|}{w/o Slow Debater} & 67.0 & 63.1 & 72.7 & 36.9 & 59.9  \\
\bottomrule
\end{tabular}
\caption{Ablation study of BELLE in terms of F1 (\%). Due to space limitation, we use the abbreviations “D1”, “D2”, “D3”, and “D4” to represent Multi-hop RAG, HotpotQA, 2WikiQA, and MuSiQue, respectively.}
\label{abalaton_study}
\end{table}

\subsubsection{Ablation Study}
In Table \ref{abalaton_study}, we select three crucial components for our ablation study.
Specifically, when we remove the question type classifier, \texttt{<Question Type>} will not be inserted into the meta prompts for the subsequent bi-level MAD system.
The first-level debate is replaced with an LLM without a debating environment, and the viewpoints are directly optimized by the second-level debate.
When we remove the second-level debate, the overall system degrades to a basic MAD system associated with question types.
The results show that removing the second-level debaters has the greatest impact regardless of the LLMs used.
It indicates that this level leverages the history of debating to make reasonable operator selection opinions, compared to the basic first-level system alone.
We also find that introducing question types as prior knowledge into the MAD system is crucial for the selection of combined operators.

In the ablation experiment involving each debater, we further explore the influence of specific debaters. 
For affirmative and negative debaters, since removing a debater would disrupt the "tit for tat" atmosphere, we maintain the number of agents unchanged by using corresponding prompts. When removing the fast debater, the modeling methods of the other debaters are also synchronously removed. To remove the slow debater, we use the last round result of the fast debater as the summary result.
We observe the following: (1) Compared to designs that completely remove the first level, using several agents of the same type at the first level to obtain operator plans is beneficial for multi-hop QA tasks. (2) Removing either the fast or slow agent adversely affects task performance to some degree, with the removal of the slow summarizer having a more significant impact.

\begin{figure}[!t]
  \centering
  \includegraphics[width=8cm, height=8cm]{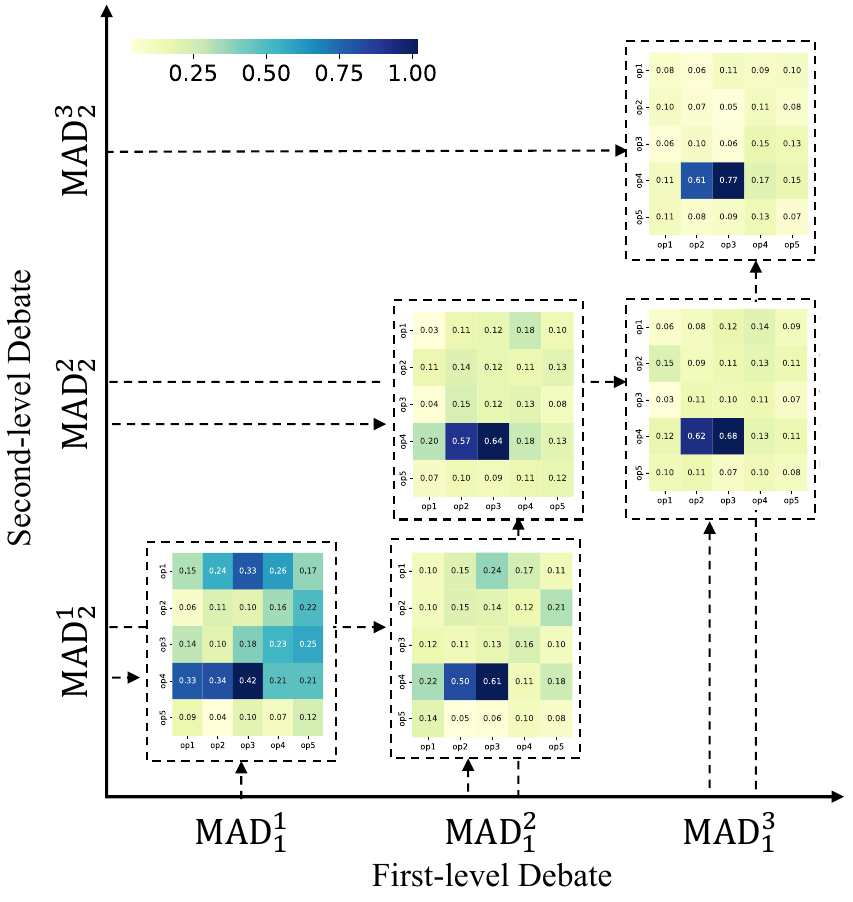}
\caption{Changes in the selection of combined operators. $\mathrm{MAD}_i^j$ denotes the debate stage at $i$-level and $j$-th debate round. (Best viewed in color.)}
  \label{heat_plot}
\end{figure}

\subsection{Detailed Analysis}
Due to space constraints, we present other detailed statistical results of our bi-level MAD system in Appendix~\ref{statistics_debaters}.

\subsubsection{Changes in Operator Selection}
From Fig.~\ref{heat_plot}, we investigate the impact of the debating contents between the first-level and second-level debaters using HotpotQA questions with the \texttt{Inference} type. Specifically, for the four important debaters in two levels, there are two situations to be considered: (1) In the same round of debating, the impact of the first-level (i.e., affirmative and negative debaters) on the second layer (i.e., slow and fast debaters) and (2) In different rounds of debating, the impact of the previous second-level on the current first-layer debating. Hence, we define the following formula to quantitatively measure the attitude change of the bi-level system:
\[ F_{f^t \rightarrow s^t}= \alpha(F_{ad}^t+F_{nd}^t) + (1-\alpha)(F_{fast}^t+F_{slow}^t) \]
and
\[ F_{s^{t-1} \rightarrow f^t} = \beta (F_{ad}^t+F_{nd}^t) + (1-\beta)(F_{fast}^{t-1}+F_{slow}^{t-1}) \]
where $F_{f^t \rightarrow s^t}$ denotes the score for situation (1) and $F_{s^{t-1} \rightarrow f^t}$ for situation (2).
Each score is a $\mathbb{R}^{5 \times 5}$ matrix, representing the combined score between 5 operators.
$F_{ad}^t$, $F_{nd}^t$, $F_{fast}^t$, and $F_{slow}^t$ represent the $t$-th round score of the four debaters, respectively.
Considering that the content discussed by the first-layer debaters in situation (1) provides information for subsequent discussion, its importance is higher.
Thus, we have assigned a value of 0.8 to $\alpha$ and 0.8 to $\beta$.
The specific score for each debater (e.g., $F_{ad}^t$) is based on the viewpoint similarity between the two operators.
We use GPT-4~\citep{gpt4_} to score the output content of debaters and the template content composed of two operators.\footnote{We define the similarity level with a corresponding score between them, such as "very similar" $\rightarrow$ 0.7.}
As shown in Fig.~\ref{heat_plot}, we observe that: (1) The bi-level MAD system becomes increasingly focused on which combined operators to use. The scores in the subgraph may fluctuate slightly, but the scoring trend of the combined operators is stable. (2) In our bi-level MAD system, the number of debate rounds is relatively small, reducing the cost of computational resources. It typically requires only 2 rounds to determine operators.


\begin{table}[t]
\centering
\scriptsize
\begin{tabular}{lccccc}
\toprule
Model $\downarrow$ Dataset $\rightarrow$ & \textbf{D1} & \textbf{D2} & \textbf{D3} & \textbf{D4} & \textbf{Avg.} \\
\midrule
\rowcolor[HTML]{C0C0C0}\multicolumn{6}{c}{\texttt{Agent-based Methods}}   \\ \midrule
\multicolumn{1}{c|}{\textbf{BELLE}} & 18,324 & 19,520 & 21,402 & 23,723 & 20,742 \\
\multicolumn{1}{c|}{LONGA.} &38,943 & 74,216 & 44,283 & 36,529 & 48,493 \\ 
\multicolumn{1}{c|}{GEAR} & 32,077 & 58,541 & 41,976 & 35,128 & 41,931 \\ 
\multicolumn{1}{c|}{RopMura} & 32,885 & 113,183 & 46,821 & 34,547 & 56,859 \\ 
\midrule
\rowcolor[HTML]{C0C0C0}\multicolumn{6}{c}{\texttt{Debate Levels}}   \\ \midrule
\multicolumn{1}{c|}{L0} & 21,376 & 26,801 & 27,542 & 26,634 & 25,588 \\ 
\multicolumn{1}{c|}{L1} & 20,988 & 24,572 & 23,894 & 27,149 & 24,151 \\
\multicolumn{1}{c|}{\textbf{L2}} & 18,324 & 19,520 & 21,402 & 23,723 & 20,742 \\
\multicolumn{1}{c|}{L3} & 23,729 & 25,863 & 31,154 & 27,269 & 27,004 \\
\midrule
\rowcolor[HTML]{C0C0C0}\multicolumn{6}{c}{\texttt{Num. of Debaters}}   \\ \midrule
\multicolumn{1}{c|}{$\mathbf{N_{f2} \rightarrow N_{s2}}$} & 18,324 & 19,520 & 21,402 & 23,723 & 20,742 \\
\multicolumn{1}{c|}{$N_{f3} \rightarrow N_{s3}$} & 26,465 & 32,841 & 28,072 & 35,917 & 30,824 \\
\multicolumn{1}{c|}{$N_{f4} \rightarrow N_{s4}$} & 32,053 & 38,716 & 34,579 & 41,839 & 36,797 \\
\multicolumn{1}{c|}{$N_{f5} \rightarrow N_{s5}$} & 39,236 & 45,170 & 42,585 & 47,736 & 43,682 \\
\bottomrule
\end{tabular}
\caption{Consumption of prompt token quantity under different agent settings. $N_{fi} \rightarrow N_{sj}$ refers to $i$ debaters in the first layer and $j$ debaters in the second layer. $L_i$ indicates different settings of the meta prompt.}
\label{computional_overhead_mad}
\end{table}

\subsubsection{Analysis of Computational Overhead}
\label{reasoning_cost}

\begin{figure}[!t]
  \centering
  \includegraphics[width=8cm,height=5.5cm]{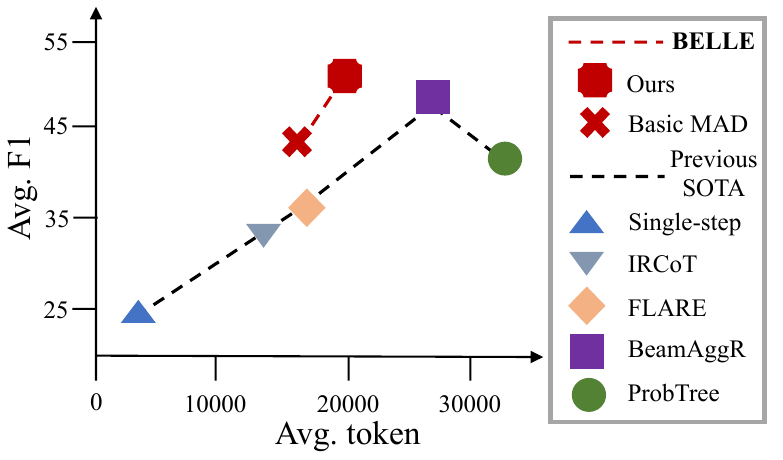}
\caption{Analysis of the relation between performance and number of retrieval tokens (Best viewed in color).}
  \label{efficiency_plot}
\end{figure}

\noindent\textbf{Comparison with Retrieval Methods.} Retrieval methods often involve frequent invocation of LLMs, resulting in significant computational overhead.
We specifically select more challenging examples of prediction errors by plain LLMs to evaluate the models. 
In Fig.~\ref{efficiency_plot}, previous methods exacerbate reasoning overhead while improving performance. In contrast, our method not only surpasses the SOTA (e.g., BeamAggR~\cite{DBLP:conf/acl/ChuCCWZDYLQ24}) in performance but also reduces reasoning overhead in terms of required tokens.
The main advantage of our model is in fully utilizing the current state and historical information, making the execution planning of the combined operators for the multi-hop question more reasonable. 
Hence, it reduces the number of rounds of combined operator retrieval and lowers the cost of prompt inference length.
Detailed statistics are in Appendix~\ref{cost_consumption}.

\noindent\textbf{Comparison with Agent-based Settings.} We compare the different debater settings, including agent-based methods, debater levels, and the number of debaters for each level.
The debate levels indicate the atmosphere of the debate prompts, as shown in Table~\ref{debate_level_table}.
As shown in Table~\ref{computional_overhead_mad}, (1) due to the establishment of a second-level reflection and judgment mechanism~\cite{DBLP:journals/corr/abs-2410-07114,DBLP:conf/nips/ZengLWLCDYXQZSL24}, our BELLE framework effectively determines the current state of the task to reduce token consumption. (2) Setting different debate levels and adjusting the number of agents for competition can improve BELLE models. By controlling the debate level of token consumption, it is unnecessary to mandate a confrontational discussion atmosphere. A relatively relaxed discussion mechanism, coupled with clear MAD system objectives, yields better results for the BELLE framework while reducing token usage. Meanwhile, excessive focus on increasing the number of agents may not necessarily enhance performance, and token consumption could increase sharply.

\section{Conclusion and Future Work}
In this paper, we introduce BELLE to effectively address the challenges of multi-hop QA by aligning specific question types with appropriate reasoning methods. 
By incorporating diverse operators and a bi-level debate mechanism, it achieves significant improvements over existing baselines.
In the future, we aim to investigate the integration of BELLE with real-world applications to assess its efficacy in dynamic and evolving environments.

\section*{Limitations}

While our proposed BELLE framework demonstrates significant improvements over existing methods, several limitations still persist. One major issue is its reliance on multiple agents interacting iteratively, especially during the debate process. Refining the debate rules and strategies could potentially reduce overhead while maintaining or even enhancing performance. Additionally, although BELLE exhibits robustness against known question types, it may struggle with novel or previously unseen question formats. To address this, adaptation to accommodate new question types will be crucial for further improvements in various applications.


\bibliography{custom}
\bibliographystyle{acl_natbib}

\appendix
\section{Implementation Details of BELLE}
\label{experimental_details}
\subsection{Model Details}
\noindent \textbf{Retrieval Setup.} To retrieve external knowledge for retrieval-augmented reasoning operators, we use the October 2017 Wikipedia dumps\footnote{\url{https://hotpotqa.github.io/wiki-readme.html}} as the candidate document pool.
Considering the computational cost of retrievers, we use the sparse model BM25 \cite{DBLP:journals/ftir/RobertsonZ09} to replace the complex models. \footnote{The retriever can be replaced by other high-precision neural models \cite{DBLP:conf/emnlp/KarpukhinOMLWEC20,DBLP:journals/tmlr/IzacardCHRBJG22} as long as the candidate documents are prepared in advance.}
We set a range of 3 to 10 candidate documents in each dataset for the multi-hop questions corresponding to these methods.

\noindent \textbf{Metrics.} The evaluation metrics are token-level EM (Exact Match), F1 and Acc (Accuracy).
The difference between EM and Acc is that EM must be strictly included in the ground-truth string, while Acc uses the LLM to perform semantic consistency checks on prediction and ground-truth.

\noindent\textbf{Baselines.} (1) \textbf{SP} denotes the standard prompting for obtaining the response. (2) \textbf{Chain-of-Thought (CoT)} generates logic reasoning steps before the final answer \cite{DBLP:conf/nips/Wei0SBIXCLZ22}. We use 4-shot for each question, providing an example for each type of question respectively. (3) \textbf{Single-step Retrieval} involves using the multi-hop question as the query to retrieve the candidate documents one time and then concatenating the search results into the prompt to perform prompt reasoning \cite{DBLP:journals/corr/abs-2203-05115}. (4) \textbf{Self-Ask} uses an iterative method to break down complex questions, progressively generating and addressing sub-questions until the final answer is reached \cite{DBLP:conf/emnlp/PressZMSSL23}. (5) \textbf{IRCoT} alternates among the retrieval-augmented reasoning methods until the retrieved information is adequate to answer the question \cite{DBLP:conf/acl/TrivediBKS23}. (6) \textbf{FLARE} dynamically adjusts the retrieval timing according to the confidence in reasoning and performs retrieval based on the subsequent reasoning sentences \cite{DBLP:conf/emnlp/JiangXGSLDYCN23}. (7) \textbf{ProbTree} breaks down the question into a tree structure, using logprobs-based aggregation of sub-questions to derive the final answer \cite{DBLP:conf/emnlp/CaoZSL0THL23}. (8) \textbf{BeamAggR} also breaks down complex questions into tree structures, which consist of atomic and composite questions, and then applies bottom-up reasoning \cite{DBLP:conf/acl/ChuCCWZDYLQ24}. (9) \textbf{EfficientRAG} iteratively generates new questions without requiring LLM calls in each round and filters out irrelevant information \cite{DBLP:conf/emnlp/ZhuangZCYLHLR0Z24}. (10) \textbf{GEAR} \cite{DBLP:journals/corr/abs-2412-18431} presents a new graph-based retriever called SyncGE, which uses an LLM to identify initial nodes for graph exploration. (11) \textbf{RopMura} \cite{wu2025talkrightspecialistsrouting} is a multi-agent system that integrates both a planner and a router to support QA across various knowledge domains.
(12) \textbf{LONGAGENT} \cite{DBLP:journals/corr/abs-2402-11550} scales LLMs (e.g., LLaMA \cite{DBLP:journals/corr/abs-2307-09288}) to a context of 128K based on MAD system and demonstrates potential superiority in long-text processing.

\noindent \textbf{Experimental Settings.} Our main experiments are conducted using GPT-3.5-turbo\cite{DBLP:conf/nips/BrownMRSKDNSSAA20} as the backbone, provided by the Azure OpenAI 2024-01-25 version.
In addition, we perform experiments using GPT-4 \cite{gpt4_}, with the Azure OpenAI 2024-06-13 version, to ensure the accuracy of classification in Sect. \ref{analysis_mhq}, despite a higher response cost.\footnote{\url{https://learn.microsoft.com/en-us/azure/ai-services/openai/}}
 To verify the effectiveness of our LLM-agnostic multi-hop QA framework, we replace the backbone of all baselines with Qwen2.5-7B \cite{qwen25} and Mistral-7B \cite{DBLP:journals/corr/abs-2310-06825}.
 
For the SFT experiment in Appendix \ref{impact_type_classifier}, we use Qwen2.5-7B-instruct, training on 8 $\times$ Nvidia A100 GPUs for about 15 hours.
We use the full tuning paradigm to perform the SFT process. The hyperparameters are as follows: batch size is 1, learning rate is 1e-5, with the AdamW optimizer \cite{DBLP:conf/iclr/LoshchilovH19}, and the number of epochs is 1.

\begin{table}[t]
\centering
\normalsize
\begin{tabular}{lcccc}
\toprule
\multicolumn{1}{c|}{\textbf{Type $\downarrow$ Data $\rightarrow$}} & \textbf{D1} & \textbf{D2} & \textbf{D3} & \textbf{D4} \\
\midrule
\multicolumn{1}{c|}{Inference} &  816 & 2158 & 4758 & 938   \\
\multicolumn{1}{c|}{Comparison} &  856 & 2495 & 3819 & 856   \\
\multicolumn{1}{c|}{Temporal} & 583 & 1033 & 2691 & 414 \\
\multicolumn{1}{c|}{Null} & 301 & 1719 & 1308 & 251  \\ \midrule
\multicolumn{1}{c|}{Total} & 2556 & 7405 & 12576 & 2459  \\ 

\bottomrule
\end{tabular}
\caption{The number of multi-hop question types included in each dataset.  ``D1'', ``D2'', ``D3'', and ``D4'' represent Multi-hop RAG, HotpotQA, 2WikiQA, and MuSiQue respectively.}
\label{main_data_statistics}
\end{table}

\subsection{Dataset Details}
\label{datasset_details}
\noindent\textbf{Datasets.} We evaluate BELLE on four open-domain multi-hop QA datasets: MultiHop-RAG \cite{DBLP:journals/corr/abs-2401-15391}, 2WikiMultiHopQA \cite{DBLP:conf/coling/HoNSA20}, HotPotQA \cite{DBLP:conf/emnlp/Yang0ZBCSM18}, and MuSiQue \cite{DBLP:journals/tacl/TrivediBKS22}. These datasets contain questions with 2 to 4 hops.
For HotPotQA, 2WikiMultiHopQA, and MuSiQue, we use the same development and test sets extracted from the original dataset similar to IRCoT \cite{DBLP:conf/acl/TrivediBKS23}.
In Table \ref{main_data_statistics}, we present the data distribution of different multi-hop question types in four datasets.
Here, we refer to the Multi-hop RAG \cite{DBLP:journals/corr/abs-2401-15391}, providing the description of different multi-hop question types as follows: (1) \textbf{Inference:} This type requires identifying the internal logical semantics of multi-hop questions and connecting them through intermediate entities for answering. The final answer is an entity string. (2) \textbf{Comparison:} This is usually achieved by comparing the similarities and differences related to the entities or topics in the multi-hop questions. The answer is typically a definitive word such as ``Yes'', ``No'' or ``Consistently''. (3) \textbf{Temporal:} These questions are mainly answered based on the sequence of events occurring at different time points. The answer is also typically words such as ``Yes'', ``No'', or a temporal indicator word like ``before''. (4) \textbf{Null:} These are questions whose answer cannot be obtained from the retrieved documents or are other free-form questions. The answer is generally a noun with an indefinite form.
Particularly, we choose the distractor setting dataset of HotpotQA \cite{DBLP:conf/emnlp/Yang0ZBCSM18}, and all hops (i.e., 2, 3, and 4-hop) in MuSiQue \cite{DBLP:journals/tacl/TrivediBKS22} are used.

\begin{figure*}[!t]
  \centering
  \includegraphics[height=7.5cm,width=15cm]{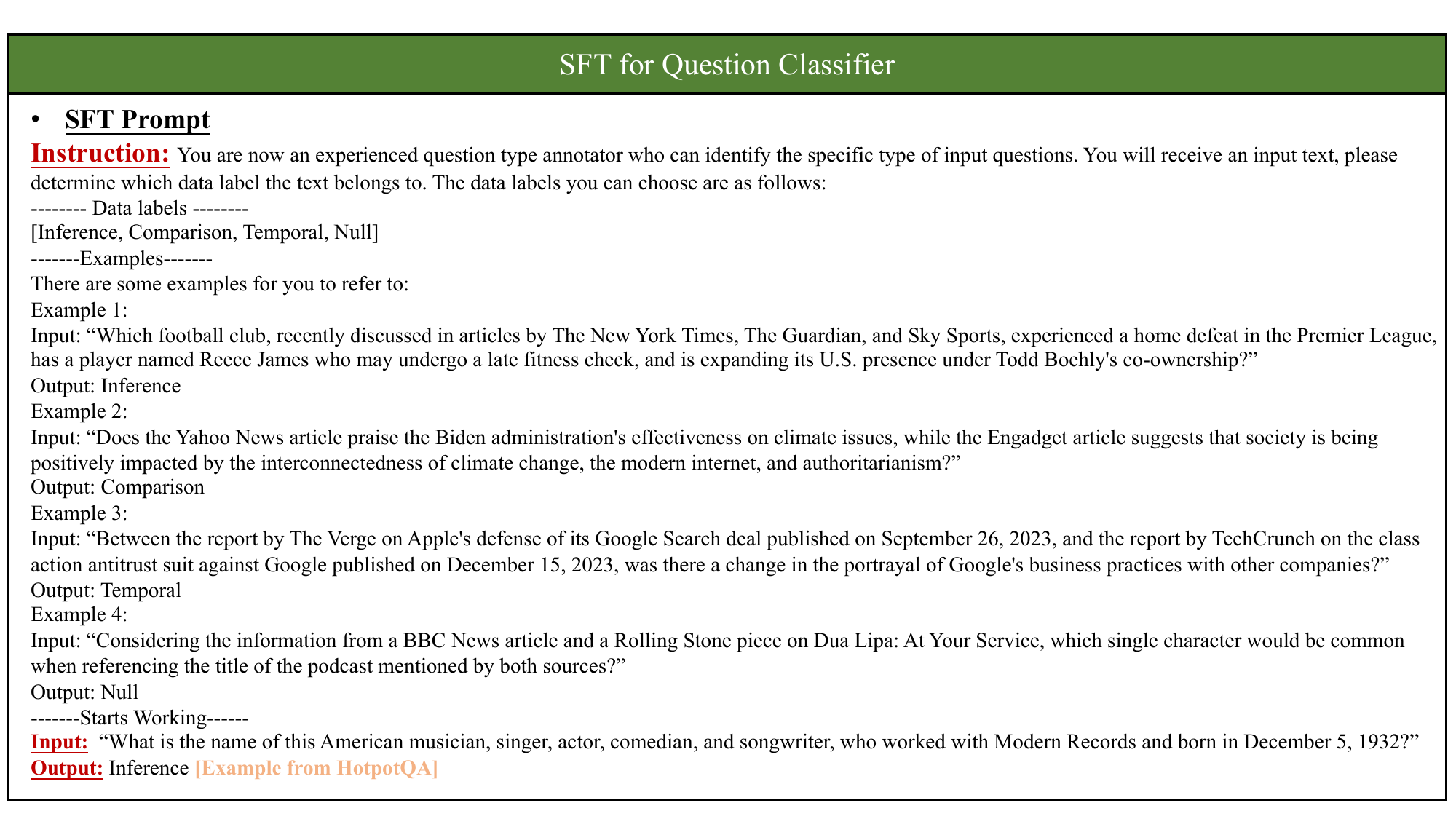}
  \caption{The SFT template for our experiment in Appendix \ref{impact_type_classifier}.}
  \label{sft_template_fig}
\end{figure*}

\noindent \textbf{SFT QA Dataset:} We collect the SFT QA pair data for the experiment of question classifier analysis in Appendix  ~\ref{impact_type_classifier}. The training prompt is shown in Fig.~\ref{sft_template_fig}. We use the training datasets of HotpotQA-hard, and 2WikiQA-hard to form the SFT data. The number of training data points is 15,661 and 12,576, respectively.

\noindent \textbf{Reasoning Cost Dataset:} To demonstrate the effectiveness and computational resource cost of our BELLE model, we design an inference consumption in Sect. ~\ref{reasoning_cost}.
We choose various retrieval-augmented reasoning methods as our strong baselines. The metrics are the retrieved tokens required and the average F1 results. We particularly select the difficult multi-hop questions as the dataset for this experiment, randomly selecting 5,000 samples with various types from the prediction errors of LLMs.

\begin{figure*}[!t]
  \centering
  \includegraphics[height=5cm,width=16cm]{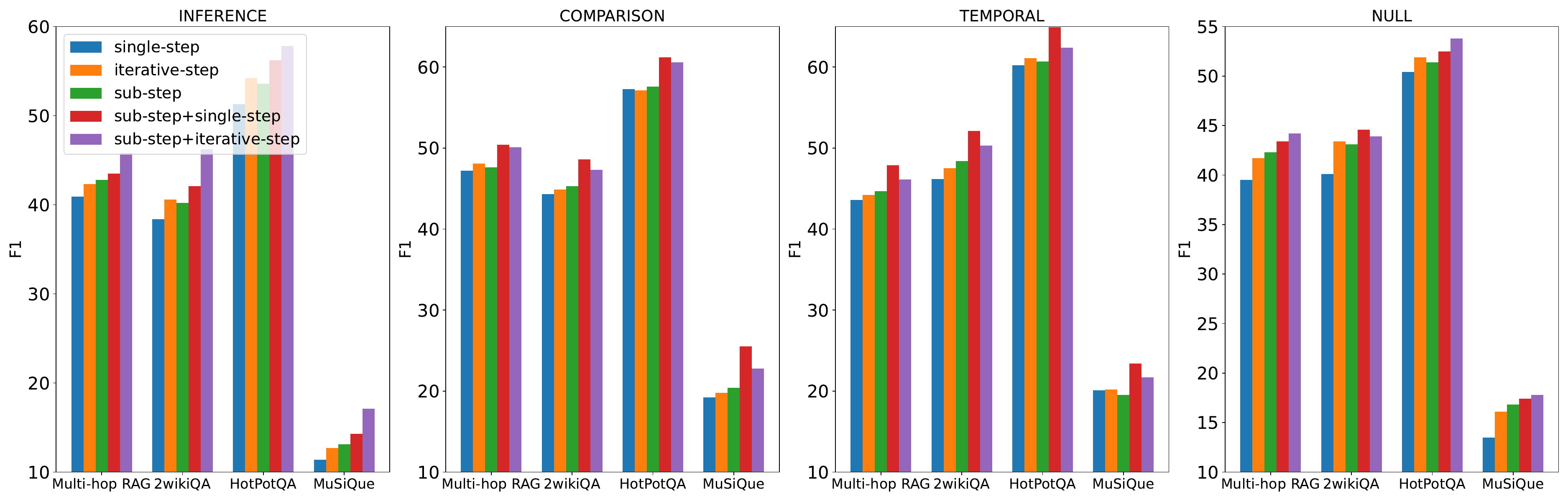}
  \caption{Overall Comparison of single and combined operators in different multi-hop questions.}
  \label{other_data_analysis}
\end{figure*}

\begin{table*}[t]
\centering
\begin{normalsize}
\begin{tabular}{ccccccccccc}
\toprule
 \multicolumn{1}{c|}{\multirow{1}{*}{Dataset$ \rightarrow$}} & \multicolumn{2}{c|}{Multi-hop RAG}                                                & \multicolumn{2}{c|}{HotpotQA}                                                     & \multicolumn{2}{c|}{2WikiQA}  & \multicolumn{2}{c|}{MuSiQue}            & \multicolumn{2}{c}{Avg.}                                             \\ \cmidrule{2-11} 
 \multicolumn{1}{c|}{\multirow{1}{*}{Models$ \downarrow$}}                       & \multicolumn{1}{c|}{\# token}   & \multicolumn{1}{c|}{F1}  & \multicolumn{1}{c|}{\#token}   & \multicolumn{1}{c|}{F1}   & \multicolumn{1}{c|}{\#token}   & \multicolumn{1}{c|}{F1}   & \multicolumn{1}{c|}{\#token}   & \multicolumn{1}{c|}{F1} & \multicolumn{1}{c|}{\#token}   & \multicolumn{1}{c}{F1}  \\ \midrule

\multicolumn{1}{c|}{Single-step}             & \multicolumn{1}{c|}{4109} & \multicolumn{1}{c|}{30.5} & \multicolumn{1}{c|}{3876} & \multicolumn{1}{c|}{29.4} & \multicolumn{1}{c|}{3652} & \multicolumn{1}{c|}{21.5} & \multicolumn{1}{c|}{4356} & \multicolumn{1}{c|}{18.3}  & \multicolumn{1}{c|}{3998} & \multicolumn{1}{c}{24.9}   \\ 

\multicolumn{1}{c|}{IRCoT}                    & \multicolumn{1}{c|}{15368} & \multicolumn{1}{c|}{39.2} & \multicolumn{1}{c|}{14677} & \multicolumn{1}{c|}{45.8} & \multicolumn{1}{c|}{13924} & \multicolumn{1}{c|}{31.6} & \multicolumn{1}{c|}{14229} & \multicolumn{1}{c|}{22.7} & \multicolumn{1}{c|}{14550} & \multicolumn{1}{c}{34.8}    \\ 

\multicolumn{1}{c|}{FLARE}       & \multicolumn{1}{c|}{17212} & \multicolumn{1}{c|}{41.4} & \multicolumn{1}{c|}{19516} & \multicolumn{1}{c|}{44.8} & \multicolumn{1}{c|}{16592} & \multicolumn{1}{c|}{33.4} & \multicolumn{1}{c|}{17285} & \multicolumn{1}{c|}{24.1} & \multicolumn{1}{c|}{17651} & \multicolumn{1}{c}{35.9}    \\ 

\multicolumn{1}{c|}{ProbTree}                 & \multicolumn{1}{c|}{30975} & \multicolumn{1}{c|}{45.7} & \multicolumn{1}{c|}{28360} & \multicolumn{1}{c|}{47.3} & \multicolumn{1}{c|}{37241} & \multicolumn{1}{c|}{37.2} & \multicolumn{1}{c|}{40032} & \multicolumn{1}{c|}{28.1}  & \multicolumn{1}{c|}{34152} & \multicolumn{1}{c}{39.6}  \\ 

\multicolumn{1}{c|}{BeamAggR}                 & \multicolumn{1}{c|}{26940} & \multicolumn{1}{c|}{{52.3}} & \multicolumn{1}{c|}{25463} & \multicolumn{1}{c|}{54.9} & \multicolumn{1}{c|}{{31943}} & \multicolumn{1}{c|}{43.6} & \multicolumn{1}{c|}{34260} & \multicolumn{1}{c|}{{30.1}}  & \multicolumn{1}{c|}{29651} & \multicolumn{1}{c}{45.2}   \\ 

\multicolumn{1}{c|}{Basic MAD}                 & \multicolumn{1}{c|}{16439} & \multicolumn{1}{c|}{{49.3}} & \multicolumn{1}{c|}{13530} & \multicolumn{1}{c|}{53.2} & \multicolumn{1}{c|}{{21402}} & \multicolumn{1}{c|}{42.5} & \multicolumn{1}{c|}{22593} & \multicolumn{1}{c|}{{28.3}}  & \multicolumn{1}{c|}{18491} & \multicolumn{1}{c}{43.3}   \\  \midrule

\multicolumn{1}{c|}{BELLE}                 & \multicolumn{1}{c|}{18324} & \multicolumn{1}{c|}{{56.4}} & \multicolumn{1}{c|}{19520} & \multicolumn{1}{c|}{62.8} & \multicolumn{1}{c|}{{22394}} & \multicolumn{1}{c|}{47.2} & \multicolumn{1}{c|}{23723} & \multicolumn{1}{c|}{{33.5}} & \multicolumn{1}{c|}{20742} & \multicolumn{1}{c}{50.0}    \\  \bottomrule
\end{tabular}
\end{normalsize}
\caption{Token consumption per multi-hop questions and performance in four datasets.}
\label{token_consumption}
\end{table*}

\begin{table}[t]
\centering
\scriptsize
\begin{tabular}{lccccc}
\toprule
Model $\downarrow$ Dataset $\rightarrow$ & \textbf{D1} & \textbf{D2} & \textbf{D3} & \textbf{D4} & \textbf{Avg.} \\
\midrule
\rowcolor[HTML]{C0C0C0}\multicolumn{6}{c}{\texttt{Qwen2.5-7B}}   \\ \midrule
\multicolumn{1}{c|}{CoT}  & 24.9 & 22.5 & 19.9 & 11.8 & 19.8 \\
\multicolumn{1}{c|}{ProbTree}  & \underline{50.7} & 47.1 & 55.6 & 17.3 & 42.7 \\
\multicolumn{1}{c|}{BeamAggR} & 55.8 & \underline{51.8} & \underline{62.4} & \underline{23.2} & \underline{48.3} \\
\multicolumn{1}{c|}{BELLE} & \textbf{64.1} & \textbf{59.4} & \textbf{68.5} & \textbf{32.8} & \textbf{56.2} \\ \midrule
\rowcolor[HTML]{C0C0C0}\multicolumn{6}{c}{\texttt{Mistral-7B}}   \\ \midrule
\multicolumn{1}{c|}{CoT} & 26.3 & 25.1 & 19.2 & 10.6 & 20.3 \\
\multicolumn{1}{c|}{ProbTree} & 51.4 & 48.7 & 53.8 & 16.9 & 42.7 \\
\multicolumn{1}{c|}{BeamAggR} & \underline{56.6} & \underline{54.3} & \underline{59.9} & \underline{22.7} & \underline{48.4} \\
\multicolumn{1}{c|}{BELLE} & \textbf{65.8} & \textbf{61.3} & \textbf{64.4} & \textbf{29.7} & \textbf{55.3} \\ \midrule
\rowcolor[HTML]{C0C0C0}\multicolumn{6}{c}{\texttt{GPT-4}}   \\ \midrule
\multicolumn{1}{c|}{CoT} & 51.8 & 47.2 & 44.9 & 24.6 & 42.1 \\
\multicolumn{1}{c|}{ProbTree} & 62.8 & 61.5 & 68.3 & 30.5 & 55.8 \\
\multicolumn{1}{c|}{BeamAggR} & 67.6 & 63.4 & 72.7 & 36.2 & 60.0 \\
\multicolumn{1}{c|}{BELLE} & \textbf{71.3} & \textbf{66.9} & \underline{75.3} & \textbf{41.3} & \textbf{63.7} \\ \midrule
\multicolumn{1}{c|}{BELLE (GPT-3.5-turbo)} & \underline{70.4} & \underline{66.5} & \textbf{75.7} & \underline{40.6} & \underline{63.3} \\
\bottomrule
\end{tabular}
\caption{Results of different LLMs in terms of F1 (\%).}
\label{diff_llms}
\end{table}

\section{Additional Experimental Discussion}
\label{other_res}
\subsection{Annotation Process of Question Types}
\label{overall_results_da}
\noindent \textbf{The Complete Results:} Considering that there are too many combinations between operators, we limit the experiment to the two most typical combinations.
In Fig. \ref{other_data_analysis}, we present the overall results for data analysis (see Sect. \ref{analysis_mhq}).
Due to the relatively small range of MuSiQue results compared to others, we have considered space limitations and placed its results in Appedix.
The conclusions in Sect. \ref{analysis_mhq} are consistently effective.

\noindent \textbf{Analysis of Question Type Annotation:} For the question type annotation process, to ensure the accuracy of data labeling, we use the GPT-4 model rather than GPT-3.5-turbo. It has been widely adopted in many works for data labeling \cite{DBLP:journals/corr/abs-2409-09659,DBLP:conf/chi/HeHDRH24,walshe2025}.
The process of cross validation involves two NLP experts conducting separate labeling and discussing results with inconsistent cases until the error is controlled within 5\%. This mechanism of labeling from coarse-grained to fine-grained manual review is widely used in many works \cite{DBLP:conf/emnlp/RajpurkarZLL16,DBLP:conf/emnlp/JingXZ19,DBLP:conf/acl/ZhangWQYCHH21}. Therefore, after selecting reliable models and experts, the labeling results of data analysis can be trusted.
Due to the flexibility of our framework, we can directly add type descriptions in \texttt{Meta prompt} to expand fine-grained multi-hop question types. For example, we have added two new types of fine-grained inference ``Bridge-comparison'' and ``Compositional'' \cite{DBLP:conf/coling/HoNSA20}.
Specifically, we add two examples and twp multi-hop QA question type descriptions in Fig. \ref{question_type_annotation}.
\begin{itemize}
    \item The \texttt{Meta Prompt} is transformed to: ``As an assistant, ...... ‘Inference’, ‘Comparison’, ‘Temporal’ , ‘Bridge-comparison’, ‘Compositional’ and ‘Null’ ''
    \item The demonstration examples are added: ``Example 5: Why did the founder of Versus die? (Output: Compositional)'' and ``Example 6: Are both director of film FAQ: Frequently Asked Questions and director of film The Big Money from the same country? (Output: Bridge-comparison) '' 
\end{itemize}

Then we perform the experiments on two new types, our BELLE framework further improves the performances over the four datasets to ``65.1 (+0.4) / 71.2 (+0.8)'', ``59.9 (+0.7) / 67.8 (+1.3)'', ``71.4 (+1.7) / 79.3 (+3.6)'', ``30.4 (+0.2) / 41.8 (+1.2) '' in terms of EM and F1 (\%) respectively.
These results indicate that by incorporating meaningful question types for multi-hop QA tasks, our framework continues to achieve performance improvements under the bi-layer reflection mechanism guided by question types. This experiment roughly verifies the effectiveness of our BELLE framework for multi-hop QA tasks with simple extensions.

\subsection{Results on Different Backbones}
To demonstrate the generalization ability of our method to various backbones, we also conduct experiments on open-source models and those with larger parameters.
We choose Qwen2.5-7B \cite{qwen25} and Mistral-7B \cite{DBLP:journals/corr/abs-2310-06825} as our open-source backbones and GPT-4 \cite{gpt4_} as the larger closed-book model.
We report the F1 metric for these datasets and the average results over 2, 3, and 4 hops in MuSiQue.

As shown in Table \ref{diff_llms}, we observe that our BELLE model with respect to 7B open-source backbones can achieve SOTA results on all four multi-hop QA datasets compared to previous strong baselines, demonstrating its model-agnostic nature and effectiveness.
On datasets Multi-hop RAG and HotpotQA, Mistral-7B performs better than Qwen2.5-7B due to the specialized training in long context dialogue ability.
When we replace them in BELLE with a larger backbone, the performance further improves on average (+0.4\%).
Since the GPT-4 needs higher price to obtain response, we use GPT-3.5-turbo to perform the main experiments.

\begin{table}[t]
\centering
\footnotesize
\begin{tabular}{lccccc}
\toprule
\multicolumn{1}{c|}{\textbf{Type Strategy}} & \textbf{D1} & \textbf{D2} & \textbf{D3} & \textbf{D4} & \textbf{Avg.} \\
\midrule
\rowcolor[HTML]{C0C0C0}\multicolumn{6}{c}{\texttt{Qwen2.5-7B}}   \\ \midrule
\multicolumn{1}{c|}{\textbf{ICL}} & 64.1 & \textbf{59.4} & 68.5 & \textbf{32.8} & \textbf{56.2} \\ \midrule
\multicolumn{1}{c|}{SFT} & \textbf{64.5} & 58.9 & \textbf{69.1} & 31.2 & 55.9 \\
\multicolumn{1}{c|}{Zero-shot} & 61.5 & 57.2 & 66.3 & 29.7 & 53.7 \\ \midrule
\rowcolor[HTML]{C0C0C0}\multicolumn{6}{c}{\texttt{GPT-3.5-turbo}} \\ \midrule
\multicolumn{1}{c|}{\textbf{ICL}}  & 70.4 & \textbf{66.5} & 75.7 & \textbf{40.6} & \textbf{63.3} \\ \midrule
\multicolumn{1}{c|}{SFT} & \textbf{70.6} & 65.8 & \textbf{75.9} & 38.2 & 62.6 \\
\multicolumn{1}{c|}{Zero-shot} & 68.1 & 63.5 & 71.3 & 36.7 & 59.9 \\
\bottomrule
\end{tabular}
\caption{Performance of multi-hop QA tasks with different question type strategies in terms of F1 (\%).}
\label{question_type_strategy}
\end{table}

\subsection{Impact of Type Classifier}
\label{impact_type_classifier}
From the results of the ablation study in Table \ref{abalaton_study}, we can find that incorporating question types is crucial for guiding our MAD system to provide reasonable planning of combined operators. Hence, we further analyze the methods used to obtain question types: in-context learning (ICL), SFT, and zero-shot prompting. 
For the ICL mechanism, we provide a sample for each type of multi-hop question combined with instructions to form the input prompt of the LLMs. 
In addition, we use the existing question types and QA pairs to test the SFT mechanism and the training datasets are described in Appendix \ref{datasset_details}.
In zero-shot prompting, we only use the instruction and label space to prompt the LLMs.
From the results in Table \ref{question_type_strategy}, although ICL may fluctuate on some datasets compared to SFT, it can achieve the best average performance regardless of the parameter size of the LLMs.
However, zero-shot prompting results in a rapid decrease in effectiveness due to the complex reasoning required for multi-hop questions.

\subsection{Detailed Reasoning Cost Results}
\label{cost_consumption}
In Table \ref{token_consumption}, we provide the comprehensive token consumption per instance, where performance is averaged across four datasets. 
We assess the computational cost by measuring the average token usage per question. Specifically, it includes calculating the cost of prompt tokens, such as demonstrations, questions, and retrieved documents.
For iterative-step methods such as IRCoT \cite{DBLP:conf/acl/TrivediBKS23}, we have summed the number of document tokens recalled by all steps.
In our BELLE model, we count the number of recalled document tokens for the combined operators.

The main advantage of our model lies in fully utilizing the current state and historical information, making the execution planning of the combined operators for the multi-hop question more reasonable. 
Hence, it can reduce the number of rounds of combined operator retrieval and lowering the cost of prompt inference length.

\begin{table}[t]
\centering
\footnotesize
\begin{tabular}{lccccc}
\toprule
\multicolumn{1}{c|}{\textbf{\# of Debaters}} & \textbf{D1} & \textbf{D2} & \textbf{D3} & \textbf{D4} & \textbf{Avg.} \\
\midrule
\rowcolor[HTML]{C0C0C0}\multicolumn{6}{c}{\texttt{Qwen2.5-7B}}   \\ \midrule
\multicolumn{1}{c|}{\textbf{2 (Default)}} & 64.1 & \textbf{59.4} & 68.5 & \textbf{32.8} & 56.2 \\ \midrule
\multicolumn{1}{c|}{$N_{f3} \rightarrow N_{s3}$} & 63.9 & 58.7 & 68.1 & 32.3 & 55.8 \\
\multicolumn{1}{c|}{$N_{f4} \rightarrow N_{s4}$} & \textbf{64.5} & 59.2 & \textbf{68.6} & 32.7 & \textbf{56.3} \\
\multicolumn{1}{c|}{$N_{f5} \rightarrow N_{s5}$} & 63.2 & 58.4 & 67.7 & 31.8 & 55.3 \\ \midrule
\rowcolor[HTML]{C0C0C0}\multicolumn{6}{c}{\texttt{GPT-3.5-turbo}}   \\ \midrule
\multicolumn{1}{c|}{\textbf{2 (Default)}}  & 70.4 & 66.5 & \textbf{75.7} & 40.6 & 63.3 \\ \midrule
\multicolumn{1}{c|}{$N_{f3} \rightarrow N_{s3}$} & 69.8 & 66.9 & 75.2 & 39.9 & 63.0 \\
\multicolumn{1}{c|}{$N_{f4} \rightarrow N_{s4}$} & \textbf{71.2} & \textbf{67.4} & 75.5 & \textbf{41.3} & \textbf{63.9} \\
\multicolumn{1}{c|}{$N_{f5} \rightarrow N_{s5}$} & 69.4 & 65.8 & 74.9 & 39.7 & 62.5 \\
\bottomrule
\end{tabular}
\caption{Results of multi-hop QA tasks with more debaters in terms of F1 (\%). $N_{fi} \rightarrow N_{sj}$ means $i$ debaters in the first layer and $j$ debaters in the second layer.}
\label{number_debaters}
\end{table}

\begin{table}[t]
\centering
\footnotesize
\begin{tabular}{lccccc}
\toprule
\multicolumn{1}{c|}{\textbf{Debate Level}} & \textbf{D1} & \textbf{D2} & \textbf{D3} & \textbf{D4} & \textbf{Avg.} \\
\midrule
\rowcolor[HTML]{C0C0C0}\multicolumn{6}{c}{\texttt{Qwen2.5-7B}}   \\ \midrule
\multicolumn{1}{c|}{\textbf{L2 (Default)}} & \textbf{64.1} & \textbf{59.4} & 68.5 & \textbf{32.8} & \textbf{56.2} \\ \midrule
\multicolumn{1}{c|}{L0} & 63.8 & 59.1 & \textbf{68.6} & 31.5 & 55.8  \\
\multicolumn{1}{c|}{L1} & 62.6 & 57.3 & 67.8 & 29.2 & 54.2  \\
\multicolumn{1}{c|}{L3} & 61.5 & 55.8 & 67.4 & 27.4 & 53.0 \\ \midrule
\rowcolor[HTML]{C0C0C0}\multicolumn{6}{c}{\texttt{GPT-3.5-turbo}}   \\ \midrule
\multicolumn{1}{c|}{\textbf{L2 (Default)}}  & \textbf{70.4} & \textbf{66.5} & \textbf{75.7} & \textbf{40.6} & \textbf{63.3} \\ \midrule
\multicolumn{1}{c|}{L0} & 69.6 & 65.7 & 73.8 & 39.4 & 62.1 \\
\multicolumn{1}{c|}{L1} & 68.2 & 63.5 & 72.4 & 38.8 & 60.7 \\
\multicolumn{1}{c|}{L3} & 67.3 & 63.1 & 71.5 & 37.5 & 59.9 \\
\bottomrule
\end{tabular}
\caption{Performance of multi-hop QA tasks with different debate levels in terms of F1 (\%).}
\label{debate_level}
\end{table}

\begin{figure*}[!t]
  \centering
  \includegraphics[height=8cm,width=16cm]{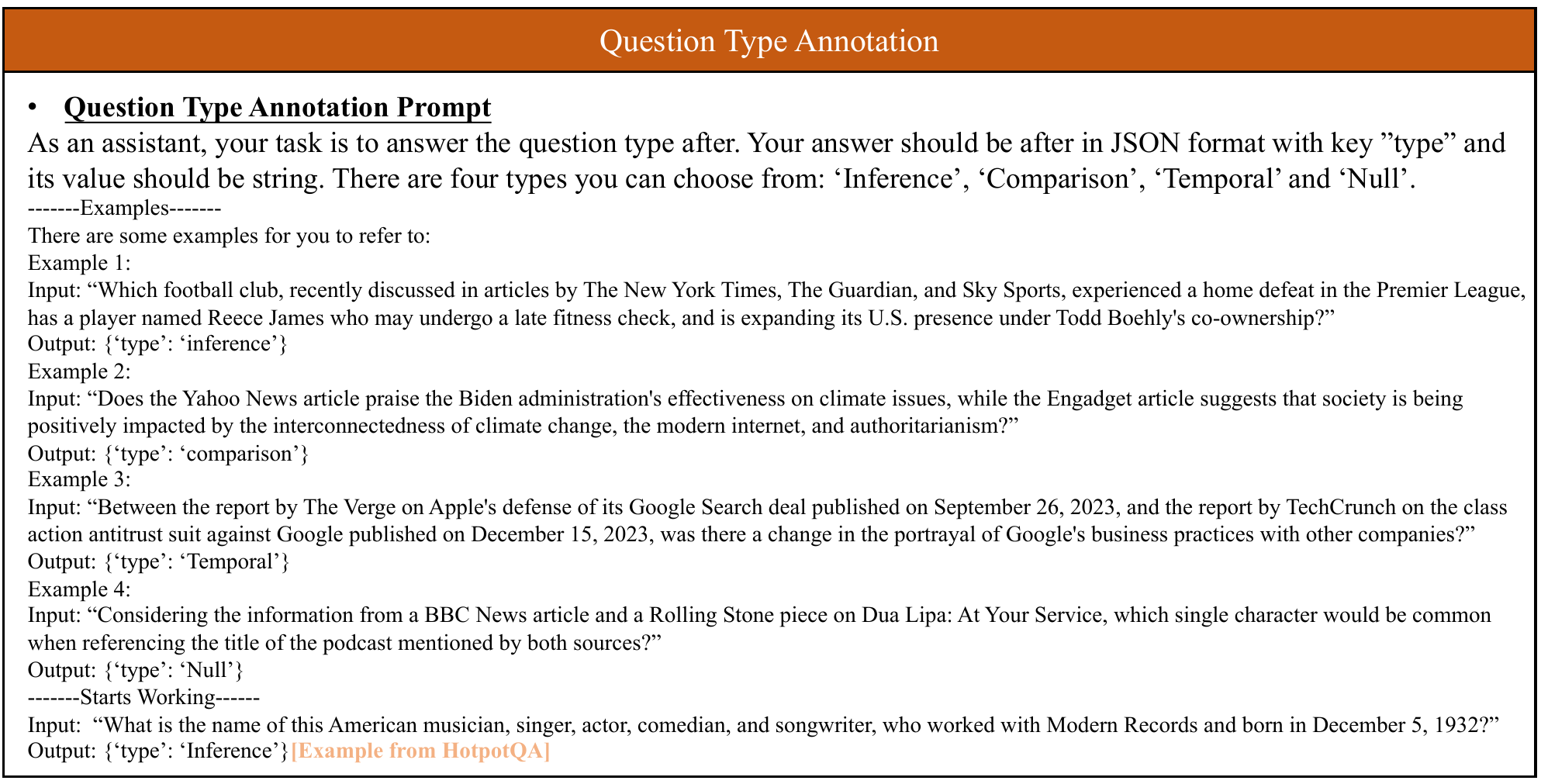}
  \caption{Prompt examples of question type annotation.}
  \label{question_type_annotation}
\end{figure*}

\subsection{Analysis of Debaters}
\label{statistics_debaters}
\noindent\textbf{Impact of Debater Number.} 
In this experiment, we increase the number of debaters in each layer for a more comprehensive discussion.
Specifically, we increase the number of debaters to three, four, and five for each layer, and then analyze the results of the bi-layer debate.
For the three debaters, we allocate two to the affirmative side and one to the negative side in the first level. The same settings apply to the second level.
We evenly allocate the number of roles within four debaters.
For the five debaters, the allocation mechanism is similar to that of three debaters.
In Table \ref{number_debaters}, we can observe that (1) As the number of debaters increases, the performance of the model decreases ($63.3 \rightarrow 63.8$ using GPT-3.5-turbo).
Considering the performance and cost of debating (see Sect. \ref{reasoning_cost}), we choose 2 debaters to report the main results.
(2) The debate effect steadily improves when the number of debaters is balanced (e.g., 2 debaters and 4 debaters).

\noindent\textbf{Impact of Debate Level.} We then study whether the atmosphere of the debate prompt has an impact on the results.
Hence, we design different instructions (see Appendix \ref{debate_levels}) to initialize the debaters' meta prompt.
In Table \ref{debate_level}, asking debaters to ``tit for tat'' is necessary for our bi-level MAD system to achieve good performance.
However, we find that ``must disagree with each other on every point'' does not lead to the best performance and may even result in a certain decrease (e.g., $\downarrow$ 3.4 in L3).
We speculate that both levels can basically reach a mutually agreed viewpoint in the early rounds of debate round friendly (see Fig. \ref{heat_plot}).

\begin{figure*}[!t]
  \centering
  \includegraphics[height=9cm,width=16cm]{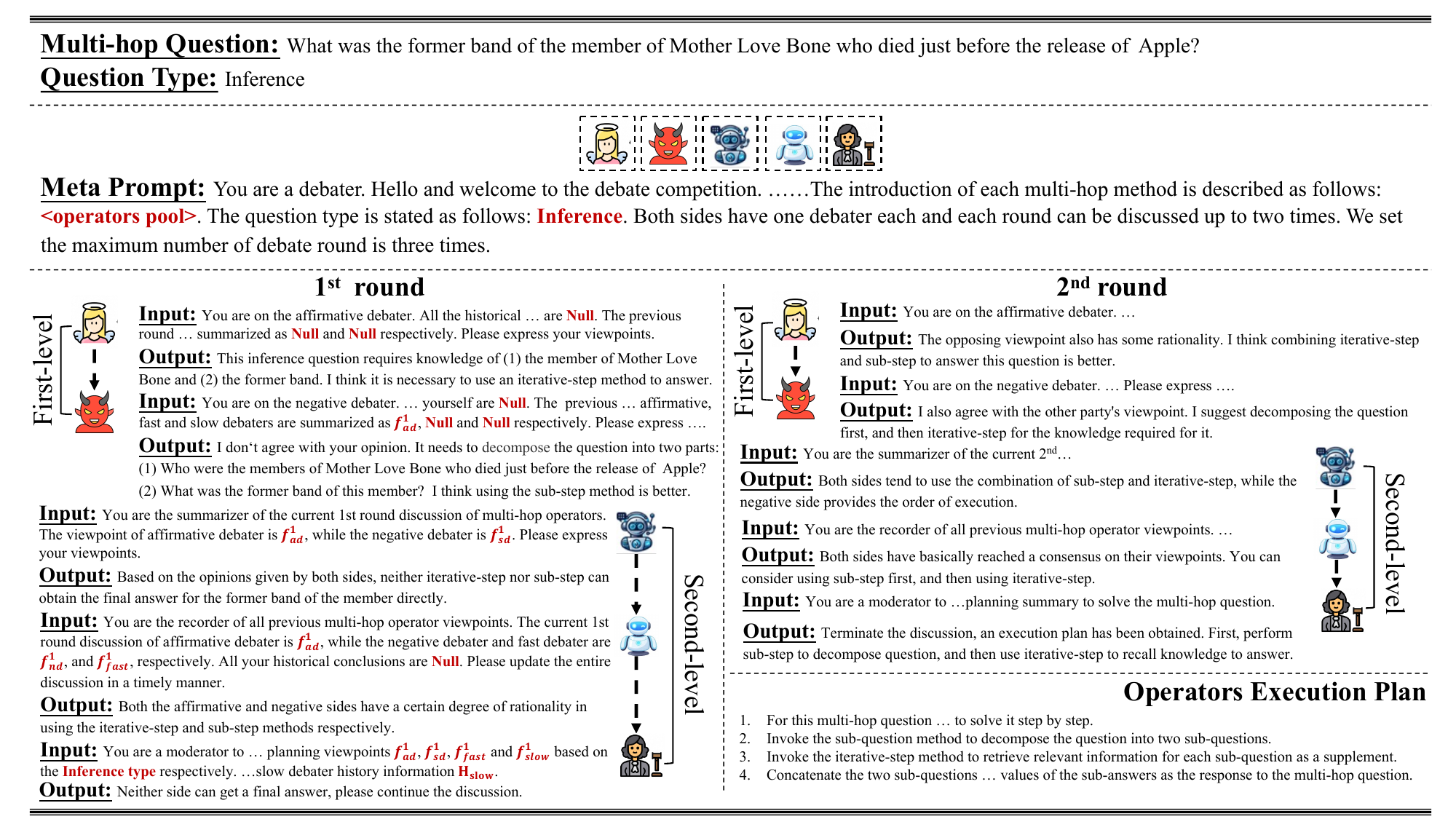}
  \caption{An example of our bi-level MAD process. Due to the excessive output content of the debater, we have replaced it with the corresponding mathematical symbols described in Sect. \ref{methodology}. In first round, we represent it using ``Null'' as some placeholder information has not been obtained yet.}
  \label{example_mad}
\end{figure*}

\subsection{Discussion of Framework Dependence}
As for the dependence of predefined heuristics and manual annotations of our BELLE framework, the previous MAD system \cite{DBLP:conf/acl/FengS00BT24,DBLP:conf/emnlp/XiongSZWWWLPL24,DBLP:conf/emnlp/Liang0JW00Y0T24} for solving NLP tasks utilizes task characteristics for prompt settings and the multi-agent collaboration design.
For the edge cases or evolving domains, the fast-debater of the second-layer judges the current discussion of the first-layer based on specific tasks without large-scale heuristic prompt debugging using \texttt{Meta Prompt}, while the slow debater comprehensively outputs a response based on historical information.
For some special task examples of edge cases or evolving domains, our second-layer MAD mechanism can perform reflective collaboration to further alleviate the possible operator viewpoint bias in high-difficulty examples at parameter scales such as GPT-3.5-turbo (e.g. 1st round to 2nd round in. Fig. \ref{example_mad}).

\section{The Templates of BELLE}
\subsection{Question Type Annotation}
\label{question_type_annotate}
Our question type annotation prompt is shown in Fig.~\ref{question_type_annotation}.
We choose an example from the HotpotQA dataset \cite{DBLP:conf/emnlp/Yang0ZBCSM18} and use GPT-4 \cite{gpt4_} to annotate the type of answer as ``\{``type'': ``Inference''\}''.
This template is also used for the question type classifier (see Sect.~\ref{question_type_classifier}), replaced with GPT-3.5-turbo \cite{DBLP:conf/nips/BrownMRSKDNSSAA20} due to the high cost of responses.

\begin{table*}[t]
\centering
\begin{scriptsize}
\begin{tabular}{cl}
\toprule
 \multirow{5}{*}{\makecell[c]{\bf Meta Prompt}} & You are a debater. Hello and welcome to the debate competition. \underline{It's not necessary to fully agree with each other's perspectives,} \\ 
& \underline{as our objective is to find the correct execution plan of operators to answer the multi-hop question based on its type.} 
You can \\ & freely combine the methods from the operator pool to solve the task. The introduction of each multi-hop method is described as \\
& follows: <operators pool>. The question type is stated as follows: <question type>. Both sides have one debater each and \\ & each round can be discussed up to two times. We set the maximum number of debate round is three times. \\ 
\midrule
\multirow{2}{*}{\makecell[c]{\bf Affirmative Debater}} & You are on the affirmative debater. All the historical round discussion results of yourself are <$H_{ad}^{t-1}$>. The previous round state  \\ & of fast and slow debaters are summarized as <$f_{fast}^{t-1}$> and <$f_{slow}^{t-1}$> respectively.
Please express your viewpoints.\\
\midrule
\multirow{3}{*}{\makecell[c]{\bf Negative Debater }} & You are on the negative debater. You disagree with the affirmative debater's points. All the historical round discussion results of \\& yourself are <$H_{nd}^{t-1}$>. The previous round state of affirmative, fast and slow debaters are summarized as <$f_{ad}^{t}$>, <$f_{fast}^{t-1}$> and \\& <$f_{slow}^{t-1}$> respectively. Please express your viewpoints.\\
\midrule
\multirow{2}{*}{\makecell[c]{\bf Fast Debater}}  & You are the summarizer of the current $t$-th round discussion of multi-hop operators. The viewpoint of affirmative debater is <$f_{ad}^t$>, \\& while the negative debater is <$f_{nd}^t$>. Please express your viewpoints.\\
\midrule
\multirow{3}{*}{\makecell[c]{\bf Slow Debater}}   & You are the recorder of all previous multi-hop operator viewpoints. The current $t$-th round discussion of affirmative debater is \\& <$f_{ad}^t$>, while the negative  debater and fast debater are <$f_{nd}^t$> and <$f_{fast}^t$> respectively.
All your historical conclusions are \\ & <$H_{slow}^{t-1}$>. Please update the entire discussion in a timely manner.\\
\midrule
\multirow{5}{*}{\makecell[c]{\bf Judge}}   & You are a moderator to give a operator planning summary to solve the multi-hop question. There is a bi-level opposing debaters \\& involved in a debate competition at the of last round. They have already presented their operator planning viewpoints <$f_{ad}$>, \\& <$f_{nd}$>, <$f_{fast}$> and <$f_{slow}$> based on the <question type> respectively. If you can get a clear summary, you can end the  \\& discussion process of the multi-hop question after outputting. If you determine that you cannot output a summary, you can extract \\& the solution from the slow debater history information <$H_{slow}$>.
\\
\bottomrule
\end{tabular}
\end{scriptsize}
\caption{The debating prompts for all debaters in our bi-level MAD system of BELLE. Each debater needs to fill content into the symbol ``<>'' before performing the discussion process.}
\label{bi_level_MAD_prompt}
\end{table*}

\subsection{Meta Prompts}
\label{meta_prompts}
Table \ref{bi_level_MAD_prompt} illustrates our meta prompt used to initialize the debaters. 
The speaking order of the debaters is as follows: affirmative debater and negative debater in the first level, followed by fast debater and slow debater in the second level, and finally the judge in each round.

\begin{table*}[t]
\centering
\begin{scriptsize}
\begin{tabular}{cl}
\toprule
 \multirow{1}{*}{\makecell[c]{\bf Level}} & \multirow{1}{*}{\makecell[c]{\bf Prompt}} \\ \midrule
\multirow{1}{*}{\makecell[c]{\bf 0}} & Both sides must reach a full consensus on every point of the debate. Each multi-hop operator selection must be agreed upon by both sides. \\ \midrule
\multirow{2}{*}{\makecell[c]{\bf 1}} & Most of the debate should be characterized by disagreements, but there may still be a small amount of consensus on less important operators selection \\ & based on question types.\\ \midrule
\multirow{2}{*}{\makecell[c]{\bf 2 (Default)}}  & It’s not necessary to fully agree with each other’s perspectives, as our objective is to find the correct execution plan of operators to answer the \\ & multi-hop question based on its type. \\
\midrule
\multirow{1}{*}{\makecell[c]{\bf 3}}   & Both sides must disagree with each other on every point of the multi-hop QA operators debate. There should be no consensus whatsoever.\\ \bottomrule
\end{tabular}
\end{scriptsize}
\caption{The different debate levels for bi-level MAD process.}
\label{debate_level_table}
\end{table*}

\subsection{An Example of Operator Planning}
\label{examples_operators}
To facilitate the readers' understanding of the operation process of our bi-level debate system, we provide an example from the HotpotQA dataset \cite{DBLP:conf/emnlp/Yang0ZBCSM18} in Fig. \ref{example_mad}, detailing how to obtain the combined operators through a step-by-step planning process.

\subsection{Different Debate Levels}
\label{debate_levels}
In Table \ref{debate_level_table}, we set four debate-level prompts to evaluate the influence of our bi-level MAD process.

\end{document}